\documentclass{article}

\usepackage{microtype}
\usepackage{graphicx}
\usepackage{subfigure}
\usepackage{booktabs} % 

\usepackage{hyperref}

\usepackage[accepted]{icml2024}

\usepackage{amsmath}
\usepackage{amssymb}
\usepackage{mathtools}
\usepackage{amsthm}
\usepackage{multirow}

\usepackage{amsmath,amsfonts,bm}

\def\eqref#1{equation~\ref{#1}}
\def\1{\bm{1}}

\def\vb{{\bm{b}}}
\def\vc{{\bm{c}}}

\def\vs{{\bm{s}}}

\def\vx{{\bm{x}}}

\def\mW{{\bm{W}}}

\DeclareMathAlphabet{\mathsfit}{\encodingdefault}{\sfdefault}{m}{sl}
\SetMathAlphabet{\mathsfit}{bold}{\encodingdefault}{\sfdefault}{bx}{n}

\newcommand{\R}{\mathbb{R}}

\usepackage[capitalize,noabbrev]{cleveref}

\theoremstyle{plain}

\theoremstyle{definition}

\theoremstyle{remark}

\usepackage[textsize=tiny]{todonotes}

\icmltitlerunning{Exploring the Benefit of Activation Sparsity in Pre-training}

\begin{document}

\twocolumn[
\icmltitle{Exploring the Benefit of Activation Sparsity in Pre-training}

\icmlsetsymbol{equal}{*}

\begin{icmlauthorlist}
\icmlauthor{Zhengyan Zhang}{thu}
\icmlauthor{Chaojun Xiao}{thu}
\icmlauthor{Qiujieli Qin}{thu}
\icmlauthor{Yankai Lin}{ruc}
\icmlauthor{Zhiyuan Zeng}{thu}
\icmlauthor{Xu Han}{thu}
\icmlauthor{Zhiyuan Liu}{thu}
\icmlauthor{Ruobing Xie}{tencent}
\icmlauthor{Maosong Sun}{thu,js}
\icmlauthor{Jie Zhou}{tencent}
\end{icmlauthorlist}

\icmlaffiliation{thu}{NLP Group, DCST, IAI, BNRIST, Tsinghua University}
\icmlaffiliation{tencent}{Tencent}
\icmlaffiliation{ruc}{Gaoling School of Artificial Intelligence, Renmin University of China}
\icmlaffiliation{js}{Jiangsu Collaborative Innovation Center for Language Ability, Xuzhou, China}

\icmlcorrespondingauthor{Zhengyan Zhang}{zy-z19@mails.tsinghua.edu.cn}
\icmlcorrespondingauthor{Maosong Sun}{sms@tsinghua.edu.cn}

\icmlkeywords{Machine Learning, ICML}

\vskip 0.3in
]

\printAffiliationsAndNotice{}  % 

\begin{abstract}
Pre-trained Transformers inherently possess the characteristic of sparse activation, where only a small fraction of the neurons are activated for each token.
While sparse activation has been explored through post-training methods, its potential in pre-training remains untapped.
In this work, we first study how activation properties change during pre-training.
Our examination reveals that Transformers exhibit sparse activation throughout the majority of the pre-training process while the activation correlation keeps evolving as training progresses. % 
Leveraging this observation, we propose Switchable Sparse-Dense Learning~(SSD).
SSD adaptively switches between the Mixtures-of-Experts~(MoE) based sparse training and the conventional dense training during the pre-training process, leveraging the efficiency of sparse training and avoiding the static activation correlation of sparse training.
Compared to dense training, SSD achieves comparable performance with identical model size and reduces pre-training costs. % 
Moreover, the models trained with SSD can be directly used as MoE models for sparse inference and achieve the same performance as dense models with up to $2\times$ faster inference speed. % 
Codes are available at \url{https://github.com/thunlp/moefication}.
\end{abstract}

\section{Introduction}

Recent studies have uncovered a notable characteristic of pre-trained Transformers: the sparse activation of neurons in their intermediate layers~\cite{zhang2022moefication,li2023the,dejavu,dong2023towards,mirzadeh2023relu}.
During inference, it has been observed that only a small fraction of the intermediate hidden states are activated, rendering a non-zero state, while the majority remain inactive.
Sparse activation presents a promising direction for improving the efficiency of Transformer-based models.

Previous work has primarily focused on leveraging the sparse activation phenomenon to speed up the inference process through post-training methods~\cite{dejavu,flash}.
For instance, with model parameters frozen, DejaVu~\cite{dejavu} proposes to selectively engage a subset of neurons likely to activate during inference, thereby reducing both the parameter communication and model computation costs.
However, the potential of utilizing sparse activation in pre-training remains largely unexplored.

Unlike the widely explored domain of post-training, where model parameters are fixed, the pre-training of Transformers is dynamic, requiring ongoing updates to model parameters.
Therefore, a preliminary step is to investigate the activation of Transformers during pre-training.
We conduct experiments on three representative text models, including GPT~\cite{GPT-2}, BERT~\cite{BERT}, and T5~\cite{T5}, with different architectures and pre-training objectives.

Our findings reveal that these models become sparsely activated in the early stage of pre-training, subsequently stabilizing in this sparse state.
It suggests that sparse activation is a pervasive phenomenon across pre-trained models, existing throughout the majority of the pre-training process.
Meanwhile, although the activation sparsity becomes stable after a certain stage of pre-training, the activation pattern is still dynamic: the set of activated neurons for a certain input varies across different stages of pre-training.
Consequently, the sparse training method for pre-training should be adaptive to the change in the activation patterns.

Based on these observations, we propose Switchable Sparse-Dense Learning~(SSD), utilizing the phenomenon of sparse activation to accelerate the pre-training of Transformers and enhance the efficiency of inference.
SSD contains two kinds of training phases: the original dense training, which facilitates the evolution of activation patterns, and the subsequent sparse training, which aims to efficiently optimize model parameters after the activation patterns have stabilized.
SSD switches between these two phases throughout the pre-training process. % 
Specifically, when the activation sparsity grows high and the activation patterns become stable, we switch to sparse training by converting the dense model to a Sparsely-activated Mixture-of-Experts~(SMoE) model, thereby enabling an efficient approximation of the original dense model.
Unlike traditional Transformers, SMoE models replace the feed-forward networks with SMoE layers, where each expert is a feed-forward network and the SMoE layer selectively engages a subset of experts, promoting computational efficiency.
To ensure the model sustains its capability for dense computation and fully leverages the model capacity, we alternate between sparse and dense training multiple times, as opposed to a permanent shift to sparse training.
This strategy aims to mitigate the risk of overfitting to the sparse computation paradigm shown in previous work~\cite{THOR}.
Besides, the final dense model is familiar with the sparse computation form, which is beneficial for the subsequent sparse inference.

In the experiments, we evaluate GPT, BERT, and T5 on language modeling and several representative downstream tasks, including sentence classification~\cite{socher2013recursive}, natural language inference~\cite{snli,mnli}, reading comprehension~\cite{rajpurkar2016squad}, and instruction-tuning~\cite{UnnaturalInstructions,supernaturalinstructions}.
Compared to traditional dense training, SSD achieves comparable performance with the same model size and fewer pre-training costs, up to $1.44\times$ speedup in FLOPs.
Besides, the models pre-trained with SSD can be used as an SMoE model for inference without any additional training, and reduces the inference time of feed-forward networks by up to $2\times$ while maintaining the performance as good as densely pre-trained models.
Moreover, by flexibly adjusting the number of selected experts during inference, our method achieves the best trade-off between performance and efficiency compared to other baseline methods, which is impossible for the models pre-trained with SMoE.

\section{Related Work}

\textbf{Activation Sparsity of Transformers.}
While non-linear activation functions are prevalent in neural networks, the activation of neurons is typically dense, for instance, 44\% of zeros in convolutional neural networks~\cite{CNNAct}.
Contrastingly, pre-trained Transformers exhibit sparse activation, with over 90\% of zeros in T5~\cite{zhang2022moefication}, and similar phenomena are observed in other pre-trained models spanning both language and vision domains~\cite{dejavu,li2023the}.
This sparsity has stimulated researchers' interest in utilizing it to accelerate inference.
There are two main approaches: neuron-based acceleration~\cite{dejavu} which dynamically selects subsets of neurons likely to be activated for computation, and SMoE-based acceleration~\cite{zhang2022moefication}, which first groups the neurons into experts and then computes in an SMoE manner by selecting the experts likely to contain most activated neurons.
The former is a fine-grained approach suitable for single-instance inference, while the latter is a coarse-grained strategy well-suited for batch inference.
In this work, we align with the SMoE-based approach, given that pre-training is commonly conducted in batch mode.

\begin{figure*}
    \centering
    \includegraphics[width=\textwidth]{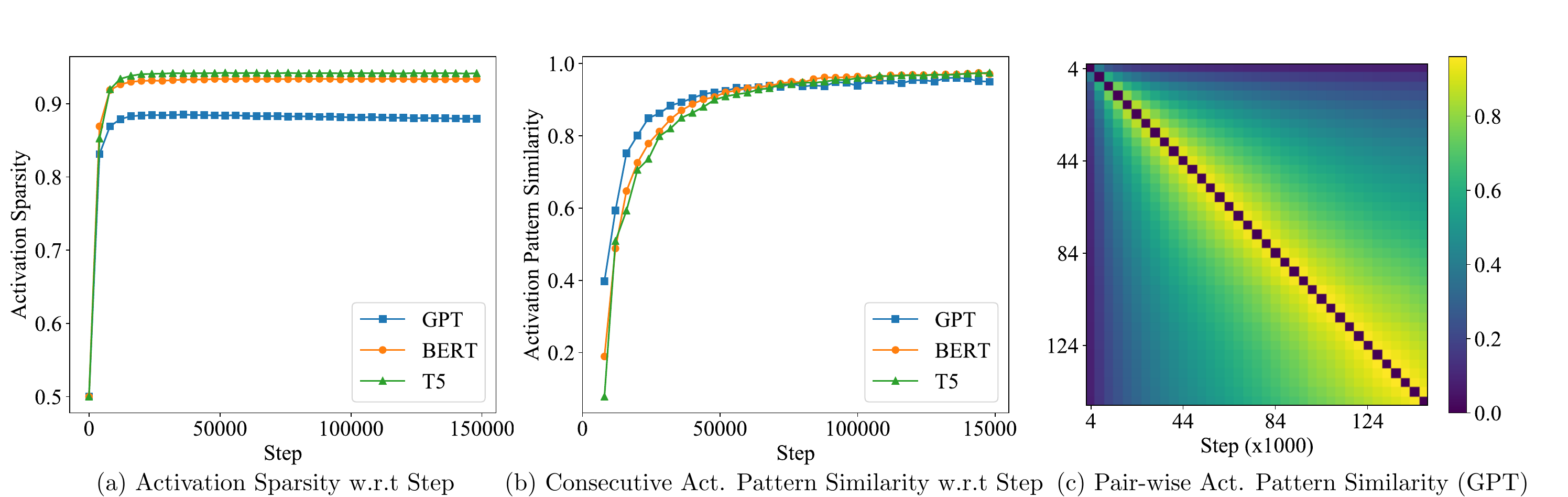}
    \vspace{-1em}
    \caption{Activation sparsity and activation pattern change of three different models during pre-training.}
    \label{fig:sparsity_structure}
\end{figure*}

\textbf{Sparse-Activated Mixture-of-Experts.} 
SMoE is a representative method to improve training efficiency of Transformer-based large language models~\cite{DBLP:conf/nips/HazimehZCSCMHC21,gao-etal-2022-parameter,zuo-etal-2022-moebert,lee-thorp-ainslie-2022-sparse,gururangan-etal-2022-demix,DBLP:conf/icml/JangKYKLLLS23,DBLP:conf/icml/0013KMH22,DBLP:conf/icml/ChenZDHLCC23,DBLP:journals/corr/abs-2306-03745} and targets both feed-forward networks~(FFNs)~\cite{baselayers,HashLayers} and attention networks~\cite{moa}.
Based on the SMoE technique, we can train Transformers that are dozens of times larger without significantly increasing computational overhead~\cite{artetxe-etal-2022-efficient,DBLP:conf/nips/RiquelmePMNJPKH21}.
However, when we evaluate the models on an equivalent parameter basis, the performance of models pre-trained with the SMoE technique frequently lags behind that of their dense counterparts~\cite{moe-dropout}.\footnote{Although Mixtral 8$\times$7B~\cite{mixtral} achieves better performance than LLaMA-70B~\cite{LLAMA2}, which is a larger dense model, the main reason may be that the pre-training corpus of Mixtral is better than that of LLaMA-70B. The direct comparison between Mixtral 8$\times$7B and LLaMA-70B is not fair.}
The performance discrepancy of certain models could potentially be attributed to a phenomenon known as representation collapse~\cite{DBLP:conf/nips/Chi0HDMPSBSMHW22}, where multiple experts redundantly encode similar information, leading to inefficient parameter utilization. 
Furthermore, the mandatory selection of all experts during inference typically does not confer any notable benefits~\cite{THOR}. 
To alleviate this issue, we combine SMoE with dense training, aiming to attain the model performance matching purely dense training while concurrently curtailing training costs.
Similar to our work, \citet{DBLP:journals/corr/abs-2404-05567} propose to add constraints during dense training to induce SMoE-like behavior thereby improving the inference efficiency of the model but it may introduce additional training overhead.

\textbf{Pre-training Acceleration Methods.}
In addition to SMoE, there are other methods to accelerate pre-training, including modifying the training objectives~\cite{electra}, inheriting the parameters from previous models~\cite{DBLP:conf/acl/ChenYS0QWWCL022,DBLP:conf/naacl/QinLYZ0ZSLLS022,DBLP:conf/icml/GongHLQWL19}, searching for appropriate hyperparameters~\cite{DBLP:conf/emnlp/IzsakBL21}, and changing the model architecture~\cite{DBLP:conf/mm/YangZLYLTXP22,PLD}.
These methods are orthogonal to our work and can be combined with our method to further improve efficiency.

\section{Preliminary Study}
\label{sec:preliminary}

We conduct a preliminary study on the evolution of the activation properties throughout the pre-training and focus on two aspects: activation sparsity and activation pattern. 
Specifically, we train three different types of PLMs, i.e., GPT, T5, and BERT, on the Pile dataset~\cite{pile}, and report the statistics of the first $150,000$ steps. More pre-training details are in Section~\ref{sec:setup}.

(1) \textbf{Activation Sparsity Change of Transformers.}
Activation sparsity is defined as the fraction of zeros in the intermediate hidden states of FFNs, which is the basis of sparse computation.
We save the model checkpoints every $4,000$ steps and calculate the activation sparsity of each checkpoint on the validation corpus.
We plot the activation sparsity of the models during training in Figure~\ref{fig:sparsity_structure}(a).
From this figure, we can see that the activation sparsity is around $0.5$ at the beginning due to the symmetry of the initialization and quickly increases to about $0.9$ after $20,000$ steps. After that, the sparsity is stable and fluctuates around $0.9$.
This observation is consistent with the previous work~\cite{mirzadeh2023relu} on auto-regressive models.
Here we extend the observation to other types of architectures and pre-training tasks, including BERT and T5.

(2) \textbf{Activation Pattern Change of Transformers.} 
Activation pattern refers to the activation correlation among neurons.
While activation sparsity stabilizes after a certain stage in pre-training, the activation pattern remains dynamic due to the ongoing updates in model parameters throughout the training process.
For instance, a pair of neurons activated together for a certain input at the onset of training may not exhibit the same behavior toward the end of training.
This dynamic nature of activation patterns poses a challenge to existing sparse acceleration approaches~\cite{dejavu}, which are primarily designed for inference and may fall short in accommodating models with significantly fluctuating activation patterns.

byHere, we study the activation pattern change of Transformers by analyzing the co-activation neuron groups.
Utilizing MoEfication~\cite{zhang2022moefication}, we categorize the neurons activated together into the same group.
By comparing the neuron groups of two checkpoints, we could measure the similarity of the activation patterns.
This grouping process essentially serves as a neuron clustering exercise, and we use the Adjusted Rand Index (ARI)~\cite{ARI} to measure the similarity between the two clustering results.
The ARI ranges from $-0.5$ to $1$, where $1$ means the two clustering results are identical and $0$ means the two clustering results are random.
We report the activation pattern similarity of consecutive checkpoints in Figure~\ref{fig:sparsity_structure}(b) and the activation pattern similarity of arbitrary checkpoints in Figure~\ref{fig:sparsity_structure}(c).

\begin{figure*}
\centering
\includegraphics[width=\textwidth]{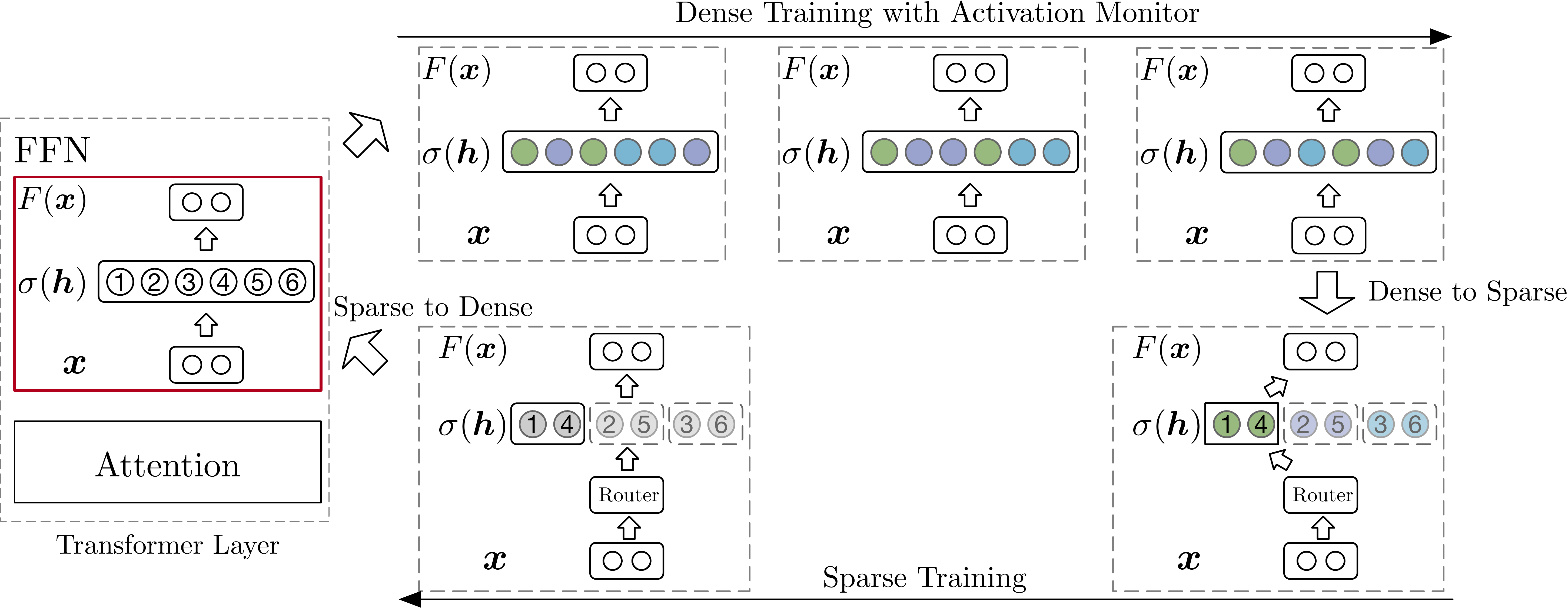}
\caption{Illustration of SSD. During dense training, we monitor the activation pattern change for each checkpoint and transform the model into an SMoE model when the activation pattern becomes stable. During sparse training, we only compute and update the parameters of selected experts for better efficiency.}
\label{fig:framework}
\end{figure*}

From these figures, we observe that the activation pattern similarity of consecutive checkpoints begins at a low point early in training, escalating to approximately $0.9$ after $50,000$ steps.
However, even as the activation pattern evolution decelerates during mid to late training stages, checkpoints separated by large step intervals continue to exhibit low activation pattern similarity.
Due to the limited computational resources, we only study the models with about 100 million parameters.
It would be interesting to investigate how the activation pattern scales with the model size in future work.

In summary, the observation of the high activation sparsity and the slow activation pattern change in the middle and late stages of training provides us with the possibility to incorporate sparse computation into dense training.
After both the activation sparsity and pattern stabilize, we can apply existing sparse acceleration methods to pre-training.
Note that the stabilization of the activation pattern unfolds at a slower pace compared to that of activation sparsity ($50,000$ steps vs. $20,000$ steps).
Consequently, we choose to employ the metric of activation pattern similarity to pinpoint the transition juncture from dense to sparse training.

\section{Method}

In this section, we first describe the overall framework of SSD and then introduce its two main components: the mechanism to transition between sparse and dense, and the criteria to determine the opportune moment for such transitions.

\subsection{Overall Framework}
In this work, we focus on accelerating the feed-forward networks within Transformers~\cite{transformer-orig}, which typically take more than 60\% of the total computation~\cite{SkillNeuron}.
The acceleration is achieved by switching between sparse and dense modes during the pre-training phase, as shown in Figure~\ref{fig:framework}.
Under sparse computation, the model is transformed into an SMoE model, incurring less computational costs compared to its original form.
The sparse activation phenomenon enables the SMoE model to emulate the original model, thus achieving a balance between efficiency and effectiveness. 
Conversely, during dense computation, all model parameters are computed and optimized to achieve better performance.
The final model reverts to a dense configuration to fully utilize the model capacity. % 
Moreover, the final model also is familiar with the sparse computation, which can be directly used for efficient sparse inference without any additional training.

In dense computation, the FFNs are computed by
\begin{equation}
\small
    \text{FFN}(\vx) = \mW_o \sigma(\mW_i \vx + \vb_i) + \vb_o,
\end{equation}
where $\mW_i \in \R^{d_{\text{ff}} \times d_{\text{model}}}$, $\mW_o \in \R^{d_{\text{model}} \times d_{\text{ff}}}$, $\vb_i \in \R^{d_{\text{ff}}}$, $\vb_o \in \R^{d_{\text{model}}}$, $\sigma$ is the activation function, and $d_{\text{ff}}$ and $d_{\text{model}}$ are the dimensions of the intermediate layer and the input/output, respectively.
For simplicity, we omit the bias term $\vb_i$ and $\vb_o$ in the following discussion.

In sparse computation, the FFNs are equally split into $N$ experts and computed in an SMoE manner,
\begin{equation}
\small
    \text{FFN}_{\text{SMoE}}(\vx) = \sum_{n=1}^N \alpha_n \mW_{o,n} \sigma(\mW_{i,n} \vx),
\end{equation}
where $\mW_{i,n} \in \R^{\frac{d_{\text{ff}}}{N} \times d_{\text{model}}}$ and $\mW_{o,n} \in \R^{d_{\text{model}} \times \frac{d_{\text{ff}}}{N}}$ are the parameters of the $n$-th expert, and $\alpha_n$ is the importance score of the $n$-th expert.
A gating network is used to score the importance of each expert for a given input $\vx$ and the experts with top-$K$ scores are selected to compute the output.
The $\alpha_n$ of unselected experts are set to $0$.
To ensure the SMoE computation is equivalent to the dense computation when $K=N$, we set the $\alpha_n$ of selected experts to $1$ through post-processing.
The details of the post-processing are provided in Appendix~\ref{sec:appendix}

\subsection{Transition between Sparse and Dense}
\label{sec:transition}

\textbf{Dense-to-Sparse Conversion.}
When the activation sparsity is high and the activation pattern is stable, we could efficiently approximate the original forward computation with sparse computation~\cite{zhang2022moefication,dejavu}.
Specifically, our approach leverages SMoE-based acceleration~\cite{zhang2022moefication} over neuron-based acceleration~\cite{dejavu} because fine-grained neuron-based selection for each token is not feasible in processing numerous tokens in a batch during pre-training.

The conversion from dense to SMoE needs to meet two requirements: (1)~the conversion should be fast to avoid additional training costs; (2)~the conversion should be smooth, ensuring the performance of the converted model remains closely aligned with the original model to avoid unstable training.
With these requirements in mind, we propose a method for fast and smooth conversion.

Specifically, the conversion contains two steps.
(1)~Neuron Clustering. 
We group the neurons that are often activated together into the same expert so that the SMoE model can efficiently compute most of the activated neurons by engaging a small fraction of experts to emulate the original model.
Inspired by~\citet{zhang2022moefication}, we cluster the rows of $\mW_i$, each of which represents a certain neuron, into $N$ groups by balanced $k$-means clustering~\cite{bkmeans}, assuming that the neurons having similar weights are more likely to be activated simultaneously.
This operation bypasses the need for directly counting the co-activation of neurons on a real-world corpus.
The counting operation is time-consuming because it requires a large number of additional forward computations and cannot be replaced by using the activation results during training due to the dynamic nature of the activation pattern.
Based on the clustering result $\vs \in \R^{d_{\text{ff}}}$, containing the corresponding expert index for each neuron, we split the weight matrices $\mW_i, \mW_o$ into $N$ sub-matrices $\mW_{i,n}, \mW_{o,n}$, respectively.
To make the conversion smoother, we propose to use the clustering results of the previous checkpoint as the initialization of clustering in the current checkpoint.
Through pilot experiments, we find that this simple strategy often provides better results, i.e., the smaller within-cluster sum of squares~(WCSS)\footnote{A metric to measure the compactness of the clustering. The smaller WCSS means the better clustering results.}, than random initialization.
The computation of WCSS is provided in Appendix~\ref{sec:appendix}.
To avert local optima, especially in early training stages where clustering may swiftly evolve, we conduct clustering twice, one with random initialization and the other with the initialization from the previous checkpoint, and select the better one.
Formally, the clustering results $\vs_j$ of the $j$-th checkpoint are computed by
\begin{equation}
    \small
    \vs_{j} = \min_{\vs \in \{f(\mW_i), f(\mW_i, \vs_{j-1})\}} \text{WCSS}(\mW_i, \vs),
\end{equation}
where $f(\mW_i)$ and $f(\mW_i, s_{j-1})$ are the clustering results with random initialization and the initialization from the previous checkpoint, respectively.
(2)~Expert Selection. We use the similarity between the input $\vx$ and the cluster centers as the importance score to select the top-$K$ experts. 
Formally, the importance score of the $n$-th expert is computed by
\begin{equation}
    \small
    \alpha_n = \vx^\top \vc_n,\quad\vc_n = \frac{N}{d_{\text{ff}}} \sum_{m=1}^{\frac{d_{\text{ff}}}{N}} \mW_{i,n}^m,
\end{equation}
where $\mW_{i,n}^m$ is the $m$-th row of $\mW_{i,n}$, and $\vc_n$ is the cluster center of the $n$-th expert.

\textbf{Sparse-to-Dense Conversion.}
The performance of SMoE models tends to lag behind their dense counterparts with equivalent parameters, primarily due to the representation collapse issue~\cite{DBLP:conf/nips/Chi0HDMPSBSMHW22,THOR}.
To optimally leverage the model capacity and avoid the overfitting of the sparse computation form, we strategically revert to dense training multiple times during training. The transition to dense is smooth given that SMoE computation aligns with dense computation when $K=N$. 
We conduct this conversion by concatenating the weight matrices of all experts, thereby obtaining the dense weight matrices, and concurrently omitting the gating network. 
This transition facilitates full-parameter optimization, effectively mitigating the representation collapse issue caused by sparse training and enabling the evolution of activation patterns.

\textbf{Discussion of Sparse Approximation.} 
Although we can approximate the model with a parameter-equivalent SMoE model, the approximation only holds for forward propagation.
If we skip the computation of some neurons during forward propagation, the gradients of these neurons will be zero during backward propagation.
However, the gradients of these neurons are usually not zero in the original model, posing an inconsistency between the SMoE model and the original model during backward propagation.
We argue that this inconsistency may not be a problem.
Intuitively, the inactivated neurons do not have strong relationships with the input, and the gradients of these neurons are not important, as the idea of Hebbian learning~\cite{seung2000half} that focuses on the neurons that are activated by the input.

\subsection{Transition Time Determination}

(1)~\textbf{Dense-to-Sparse Conversion.}
Considering the dynamic nature of the model activation during pre-training, we conduct the conversion when the activation sparsity is high and the activation pattern is stable.
Inspired by the observation in Section~\ref{sec:preliminary}, we propose to monitor the activation pattern change to determine the transition time, where the activation pattern similarity reflects the changing rate of the activation pattern.
Specifically, we set a threshold $\tau$ and switch to sparse training when the activation pattern similarity between two consecutive checkpoints is larger than $\tau$.
(2)~\textbf{Sparse-to-Dense Conversion.}
To have a controllable speed ratio, we propose to maintain a constant ratio of sparse training steps to all training steps $r$.
For example, if $r=0.5$, we will train the model with $50\%$ of the training data in the sparse training phases and the remaining $50\%$ in the dense training phases.
Specifically, we set the steps of sparse training to $T=\frac{r}{1-r}$ times the steps of the last dense training.
We give a detailed example in Figure~\ref{fig:case-loss}.
And, to ensure the final model can be used densely, we adopt dense training at the end of the training.
Hence, the pre-training process consists of several dense-sparse cycles and one final dense training.
We will pre-define the ratio of the final dense training steps as $l$, and the adjusted $T$ is calculated as $T=\frac{r}{1-r-l}$.

\section{Experiments}

\begin{figure*}
\centering
\includegraphics[width=0.85\textwidth]{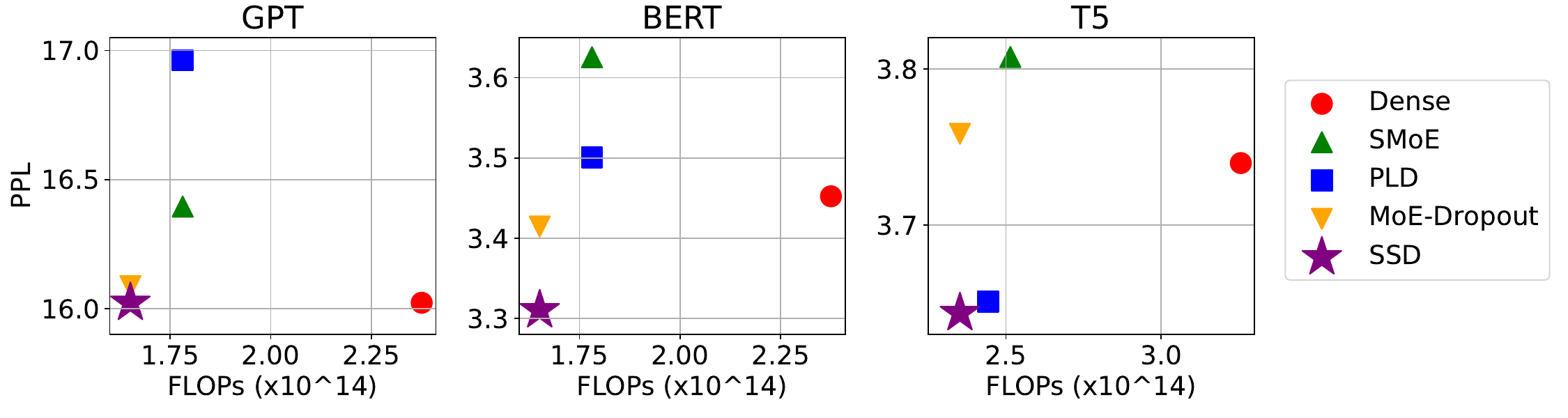}
\vspace{-1em}
\caption{Computational costs (FLOPs) of pre-training and perplexity (PPL) on the validation set of different methods on three representative models. Smaller computational cost means better efficiency and smaller perplexity means better performance.}
\label{fig:loss_plot}
\end{figure*}

\begin{figure*}
\centering
\subfigure[Perplexity w.r.t computational costs.] {
    \label{fig:variant_k_ppl}
    \includegraphics[width=0.45\linewidth]{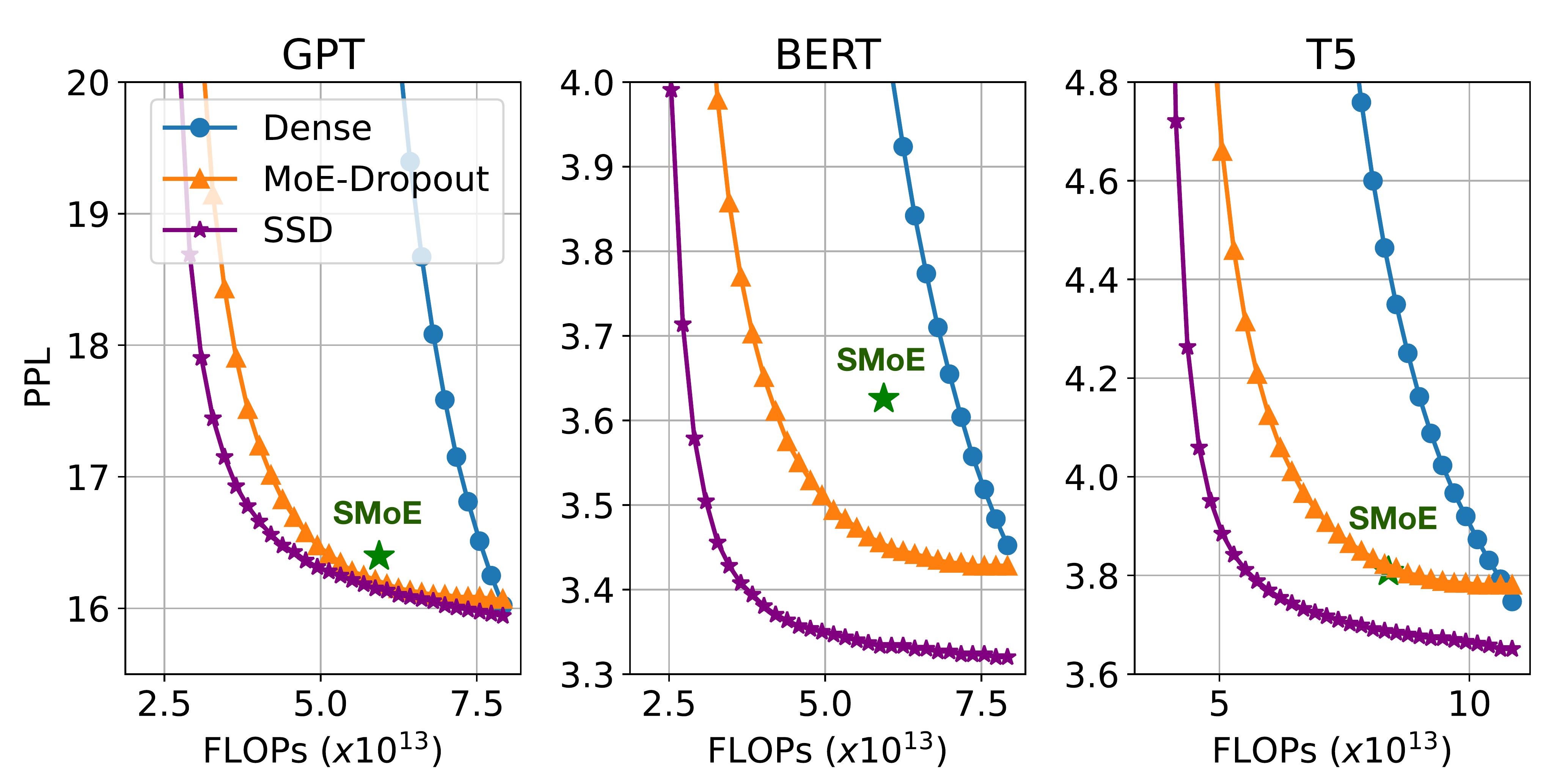}    
}
\subfigure[Perplexity w.r.t inference time.]{
\label{fig:variant_k_ppl_time}
\includegraphics[width=0.45\linewidth]{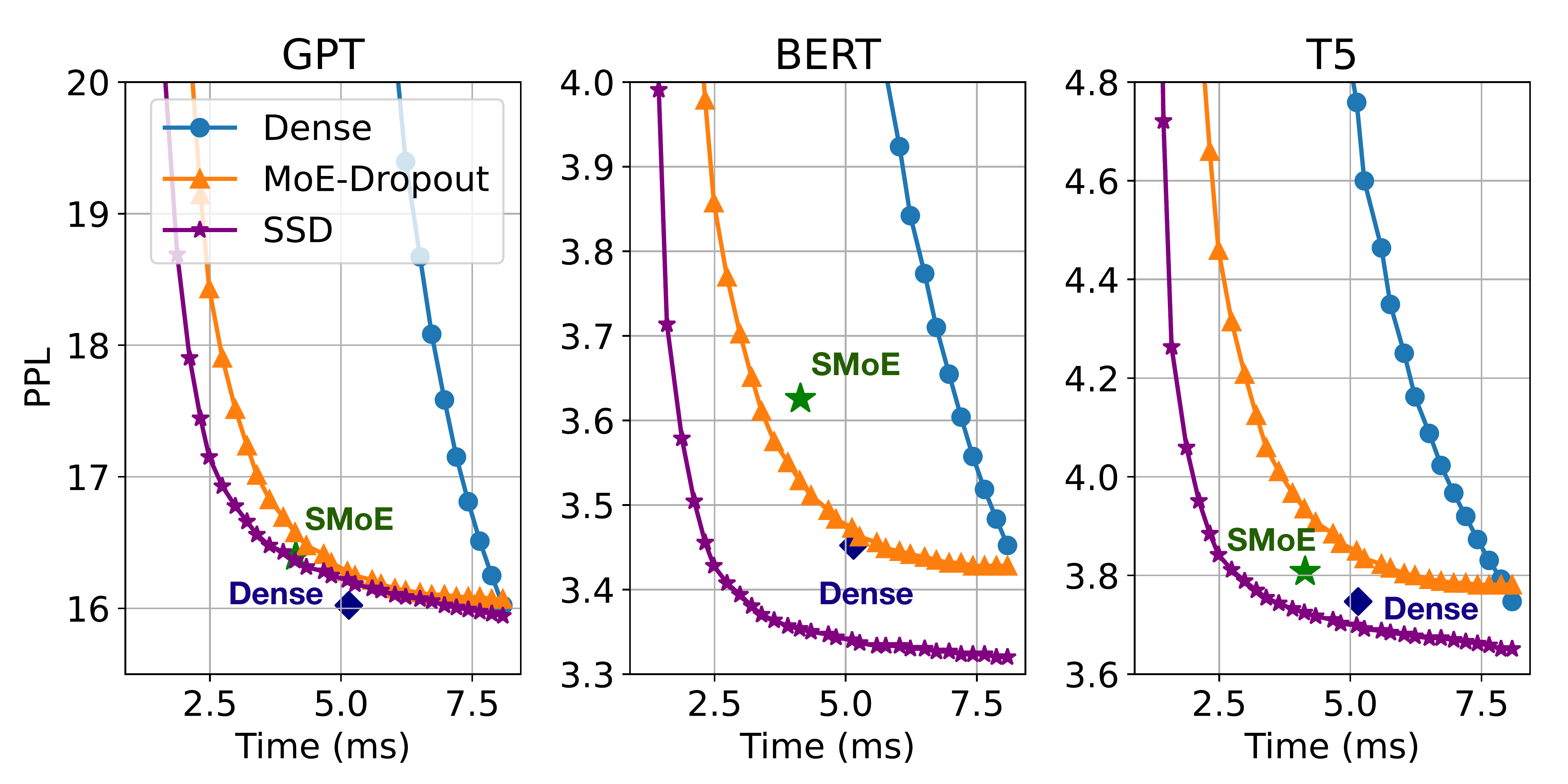}
}
\caption{Perplexity on the validation set with different computational costs and inference time by varying the number of selected experts. For comparison, we also scatter the results of SMoE with its default number of selected experts and dense models without any sparsity.}
\end{figure*}

\subsection{Experimental Setup}
\label{sec:setup}

Here we briefly introduce the setups of our experiments. Please refer to Appendix~\ref{sec:appendix} for more details.

\textbf{Baselines.} We compare our method with the following baselines: 
(1) \textit{Dense}: We compute and update all the parameters in the network, which is the default way of pre-training.
(2) \textit{SMoE}: We replace the FFNs in the \textit{Dense} baseline with MoE layers and train the model in a sparsely-activated manner~\cite{Switch-Transformer}.
Specifically, the number of parameters in experts (not including router networks) is the same as the number of parameters in the FFNs.
The final model is computed sparsely.
(3) \textit{Progressive Layer Dropping (PLD)}~\cite{PLD}: PLD randomly drops layers during training to reduce the costs.
After pre-training, we use all the layers of the final model, i.e., the same as the \textit{Dense} baseline.
(4) \textit{MoE-Dropout}~\cite{moe-dropout}:
At the beginning of pre-training, the model is an SMoE model.
During the training, MoE-Dropout gradually increases the number of selected experts $K$ to the number of experts $N$.
The final model is also densely computed. % 

\textbf{Datasets.}
(1) \textit{Pre-training corpus}: We use the Pile dataset~\cite{pile} as the pre-training corpus. Due to limited computational resources, we use the first part of the Pile dataset with over 27GB of text data.
(2) \textit{Downstream tasks}: We consider two kinds of downstream tasks, i.e., natural language understanding and instruction tuning.
More details are in Appendix~\ref{sec:appendix}.

\textbf{Training Details.}
(1) \textit{Model architecture}: In this work, we evaluate our method on all three variants of Transformers, i.e., encoder-only BERT, decoder-only GPT, and encoder-decoder T5.
For these models, we use the base version with 12 layers, 768 hidden size, and 12 attention heads for each encoder/decoder.
Following~\citet{moe-dropout}, we set the intermediate size of FFNs to 6,144.
For MoE-Dropout and SSD, we set the number of experts to 32 and the number of selected experts to 6.
For SMoE, we set the number of experts to 3 and the number of selected experts to 2 to ensure similar computational costs to MoE-Dropout and SSD.
(2) \textit{Pre-training}: 
The training epoch is set to 10, which contains about 200,000 steps, and the warmup steps are set to 2,000.
BERT and T5 adopt masked language modeling (MLM) as the pre-training task, and GPT adopts causal language modeling (CLM) as the pre-training task.
(3) \textit{Fine-tuning}: 
In this stage, we use two fine-tuning strategies for SMoE, i.e., sparse fine-tuning and dense fine-tuning, which are denoted as SMoE and SMoE (D), respectively.
For the other models, we only use dense fine-tuning.
We run each experiment 5 times and report the average of their best results on the development set.
For the instruction tuning task, which contains zero-shot transfer, we use the best checkpoint of each run on the development set for evaluation.

\subsection{Main Results}

In this subsection, we study the training efficiency and inference efficiency of different methods.

\textbf{Training Efficiency.}
We report the computational costs per batch and performance of different methods in Figure~\ref{fig:loss_plot}. 
For model performance, we evaluate perplexity on a held-out validation corpus.
From this figure, we have three observations.
(1) 
Although SMoE training can reduce the computational costs, the perplexity of SMoE is consistently higher than that of dense training.
(2) PLD and MoE-Dropout can also reduce the cost while keeping the perplexity comparable to that of dense training.
However, in some cases, the perplexity of PLD and MoE-Dropout is higher than that of dense training, such as GPT with PLD and T5 with MoE-dropout.
(3) SSD has the same speedup as MoE-dropout (up to 1.44$\times$) and achieves slightly lower or equal perplexity compared with dense training, i.e., the data points of SSD are placed at the bottom left corner of the figure.
It indicates that SSD achieves the best trade-off between training costs and performance.

\textbf{Inference Efficiency.}
We conduct experiments to investigate whether these dense models can be computed sparsely for efficient inference without additional training.
We consider three methods, i.e., Dense, MoE-Dropout, and SSD, and vary the number of selected experts $k$ from $1$ to $32$, which is the total number of experts.
For the models pre-trained with Dense, we use MoEfication to transform them into SMoE models.
For the models pre-trained with MoE-Dropout and SSD, we directly adopt their MoE structure used in pre-training.
Note that we also try to use MoEfication to transform the models pre-trained with MoE-Dropout and SSD into SMoE models, but the performance is slightly worse than that of directly using their original MoE structure.
Additionally, we also report the performance of the models pre-trained with SMoE.
We also try to vary the number of selected experts $k$ for the models pre-trained with SMoE, but the performance is consistently worse than that of using the original number of selected experts.
Therefore, we do not report the results of SMoE with different $k$.

We report the perplexity on the validation set with different computational costs per batch in Figure~\ref{fig:variant_k_ppl}.
From this figure, we have three observations.
(1) The performance of the models pre-trained with Dense is consistently worse than that of the models pre-trained with SMoE, MoE-Dropout, and SSD.
Both MoE-Dropout and SSD adopt sparse training during pre-training, which makes them more suitable for sparse inference.
It indicates that \textbf{computing sparsely during pre-training is necessary for sparse inference}.
(2) With the same computational cost, the performance of the models pre-trained with MoE-Dropout and SSD is overall better than that of the models pre-trained with SMoE.
It shows the potential of building various SMoE models with different computational costs based on a single dense model instead of training multiple SMoE models with different numbers of experts from scratch.
(3) The curve of SSD is always below that of MoE-Dropout, which indicates that SSD achieves a better trade-off between costs and performance than MoE-Dropout during inference.

We further investigate the inference time with different numbers of selected experts in Figure~\ref{fig:variant_k_ppl_time}.
Specifically, we report the inference time of a single MoE layer with different numbers of selected experts.
The inference time is measured on a single NVIDIA RTX 3090 GPU, which is a popular GPU for LLM inference, with a batch size of 32 and a sequence length of 512.
The MoE layer is implemented with the ScatterMoE library~\cite{ScatterMoE}.
From this figure, we observe that with the same inference time, the performance of the models pre-trained with SSD is still better than that of the models pre-trained with SMoE and MoE-Dropout.
Besides, since the MoE implementation is different from the original FFN implementation, the inference time of dense models without any sparsity (the diamond points with dark blue color)is shorter than that of the MoEfied dense models with large $k$.
Despite this, the performance of SSD models is still better than that of dense models without any sparsity in the cases of BERT and T5 with the same inference time.
Moreover, BERT trained with \textbf{SSD achieves the same performance as that of the dense model with less than half the inference time} (2.3ms vs. 5.1ms).
It indicates that SSD can provide competitive performance and efficiency compared to original dense models and other sparse models during inference.

\subsection{Dynamic Top-$k$}

\begin{figure}
\centering
\includegraphics[width=0.9\linewidth]{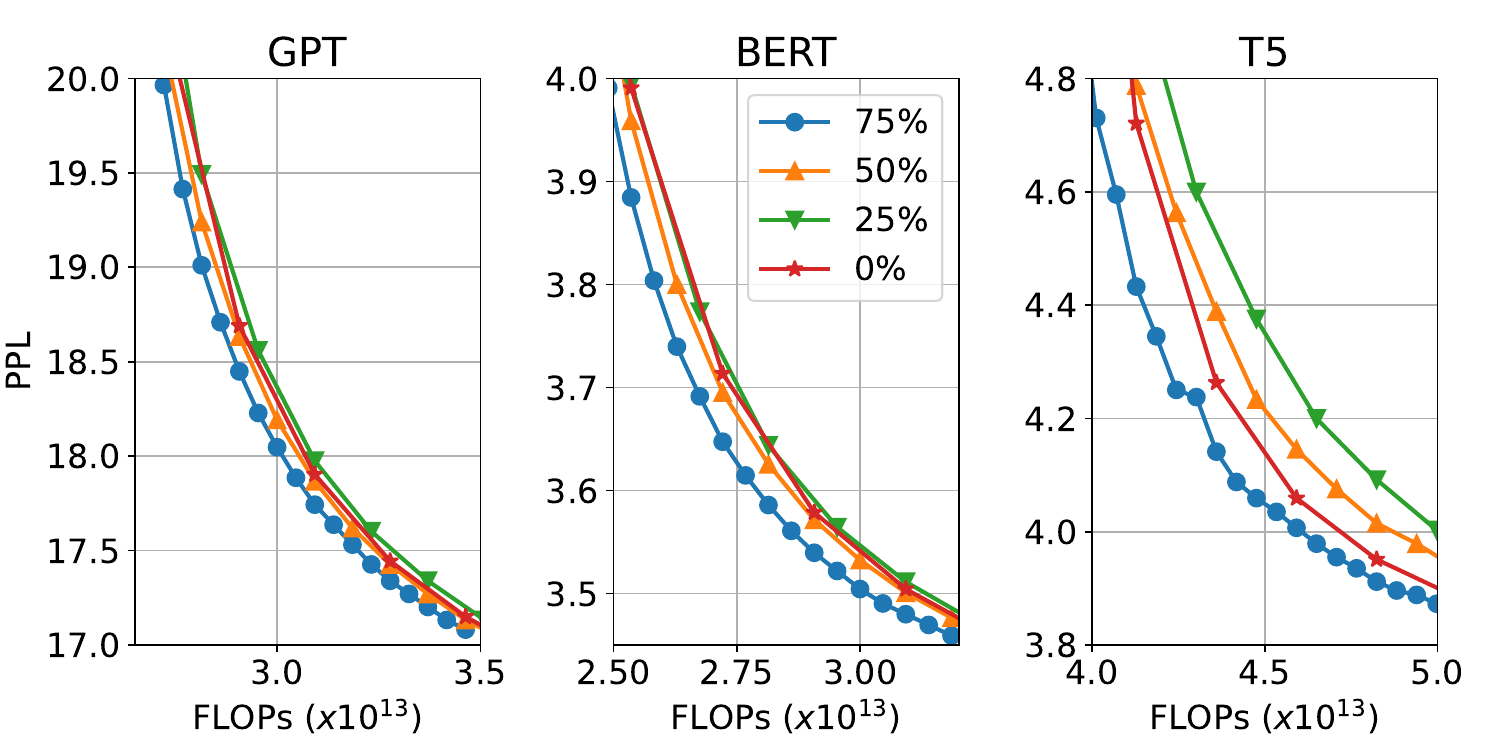}
\caption{Perplexity on the validation set with different computational costs by truncating the experts with small importance scores.}
\label{fig:vary_k_dyn}
\vspace{-1em}
\end{figure}

Based on SSD, we further investigate whether we can vary the number of selected experts for each token in an input sequence.
For example, some important tokens may need more experts to compute, and some unimportant tokens may need fewer experts to compute.
Specifically, we first compute the top-$k$ experts for each token in the input sequence as the expert candidates.
Then, we truncate the experts with small importance scores based on a given ratio.
We report the perplexity on the validation set with different truncation ratios in Figure~\ref{fig:vary_k_dyn}.
From this figure, we observe that truncating $75\%$ of the experts can consistently achieve better performance than that of using a fixed number of experts for each token under the same computational cost.
It opens up a new direction for future research to dynamically identify the number of experts for each token in an SMoE model.

\subsection{Performance on Downstream Tasks}

\begin{table}[]
\small
\centering
\caption{Evaluation results (\%) on natural language understanding tasks. MoE-D refers to MoE-Dropout. SMoE (D) refers to densely fine-tuned SMoE. Since the training costs of SMoE is smaller than other methods, we color the results of SMoE in {\color{gray} gray}.}
\label{tab:nlu}
\setlength\tabcolsep{3pt}
\begin{tabular}{lccccccc}
\toprule
& SST2 & SNLI & MNLI & QNLI & QQP & SQuAD & \multirow{2}{*}{Avg.} \\
& Acc.  & Acc.  & Acc.  & Acc  & Acc. & F1    &      \\
\midrule
\multicolumn{8}{c}{BERT-based Models} \\ \midrule
Dense &  90.0    &   88.4   &     80.1 &   87.2   &  89.7   &    78.9  & 83.2       \\ % 
{\color{gray} SMoE} & {\color{gray} 84.3} & {\color{gray} 83.9} & {\color{gray} 71.4} & {\color{gray} 82.3} & {\color{gray} 85.9} & {\color{gray} 68.8} & {\color{gray} 76.2} \\ % 
SMoE (D) & 84.7 & 84.5 & 71.8 & 82.8 & 86.5 & 70.9 & 77.2 \\ % 
PLD   &   90.0   &  88.3    &  79.5    &   86.6   &  89.8   &     77.9  & 82.8      \\ % 
MoE-D &  90.7    &   88.5   & 80.7     &  87.4    &  89.8   &       76.8  & 82.8    \\ % 
SSD   & 90.6    &  88.7    &  80.6    &  88.7    &  90.0   &     78.8  & 83.6      \\ \midrule % 
\multicolumn{8}{c}{T5-based Models} \\ \midrule
Dense &   91.5   &   89.4   &   81.7   &  88.8    &   90.2  &  83.9   & 85.9        \\ % 
{\color{gray} SMoE} & {\color{gray} 86.0} & {\color{gray} 87.5} & {\color{gray} 77.9} & {\color{gray} 84.2} & {\color{gray} 88.3} & {\color{gray} 78.5} & {\color{gray} 81.6}\\ % 
SMoE (D) & 86.2 & 87.8 & 78.6 & 84.7 & 88.6 & 78.8 & 82.0\\ % 
PLD   &   92.1   &  89.5    &   82.8   &  89.9    &  90.4   &     85.1   & 86.7     \\ % 
MoE-D &   91.8    &   89.4   &  82.3    &   89.5   &   90.4  &   83.6   & 86.0  \\ % 
SSD   &   92.5    &  89.8    &   82.5   &  89.5    &   90.5  &        85.0   & 86.6 \\ % 
\bottomrule
\end{tabular}
\end{table}

\begin{table}[]
\small
\centering
\caption{Evaluation results (\%) on instruction tuning. MoE-D refers to MoE-Dropout. SMoE (D) refers to densely fine-tuned SMoE. ``Dev'' represents the development set and ``NI'' represents Super-NaturalInstructions. Since the training costs of SMoE are smaller than other methods, we color the results of SMoE in {\color{gray} gray}.}
\label{tab:instruction}
\begin{tabular}{lccc}
\toprule
& \multicolumn{1}{c}{Dev}     & \multicolumn{1}{c}{NI}       & \multirow{2}{*}{Avg.}  \\
& \multicolumn{1}{c}{Rouge-L} & \multicolumn{1}{c}{Rouge-L} &  \\
\midrule
\multicolumn{4}{c}{GPT-based Models} \\ \midrule
Dense &    19.2     &   16.7      &  18.0 \\
{\color{gray} SMoE} & {\color{gray} 17.6} & {\color{gray} 15.9} & {\color{gray} 16.8} \\
SMoE (D) & 18.3 & 17.9 & 18.1 \\
PLD   &  19.1       &  17.5       &  18.3 \\
MoE-D &   18.8      &   18.2      &  18.5 \\
SSD   &  19.4       &    18.0    &   18.7 \\ \midrule
\multicolumn{4}{c}{T5-based Models} \\ \midrule
Dense &      19.7   &    19.1     &  19.4 \\
{\color{gray} SMoE} & {\color{gray} 16.7} & {\color{gray} 15.8} & {\color{gray} 16.3} \\
SMoE (D) & 17.0 & 16.3 & 16.7 \\
PLD   &     19.5   &    20.1     & 19.8 \\
MoE-D &      18.6   &    18.9     & 18.7 \\
SSD   &      18.6   &    20.4     &  19.5 \\
\bottomrule    
\end{tabular}
\end{table}

We report the model performance on natural language understanding tasks in Table~\ref{tab:nlu}, focusing on BERT and T5.
Besides, We report the model performance on instruction tuning in Table~\ref{tab:instruction}, focusing on GPT and T5.
The results with variance are reported in Appendix~\ref{sec:std}.
From these tables, we have three observations.
(1) The perplexity is consistent with the overall performance on downstream tasks.
For example, PLD and SSD achieve the lowest perplexity on T5, and they also achieve the top two overall performances on downstream tasks.
The same phenomenon also appears on BERT and GPT.
It indicates that the perplexity can be used as a good performance indicator on downstream tasks, which is also shown by \citet{GPT-3,gordon-etal-2020-compressing}.
(2) Densely fine-tuning SMoE can achieve better performance than sparsely fine-tuning SMoE while still being worse than other methods.
It indicates that pre-trained SMoE models cannot fully utilize the model capacity even with dense fine-tuning.
(3) SSD achieves slightly better overall performance than dense training on all evaluation settings, the only one that achieves this result among the acceleration methods.
It demonstrates the general applicability of SSD to different models and tasks.

\subsection{Scalability}

Due to the limited computational resources, we assess the scalability of SSD on large models by continual pre-training, which requires fewer training steps than pre-training from scratch.
Specifically, we continue pre-training Persimmon-8B on a diverse Chinese corpus, containing encyclopedia, news, books, and web texts.
Persimmon-8B~\cite{persimmon-8b} is a sparse-activated large language model (LLM) and has competitive performance to LLaMA-7B~\cite{LLAMA2} on several evaluation benchmarks.
We compare SSD with dense training under the same training steps with nearly 4 billion tokens.
The details of the training setup are shown in Appendix~\ref{sec:appendix}.
We evaluate the model performance on C-Eval~\cite{DBLP:journals/corr/abs-2305-08322}, a widely-used Chinese benchmark and report the average accuracy of subtasks in Table~\ref{tab:scalability}.
From this table, we observe that SSD also achieves comparable performance to dense training, which demonstrates the scalability of SSD to LLMs.

\begin{table}[h]
\centering
\caption{Performance (\%) of Persimmon-8B on C-Eval. ``Social'' represents social science, ``Human'' represents humanities, and ``Other'' represents other subjects.}
\label{tab:scalability}
\small
\begin{tabular}{lccccc}
\toprule
            & STEM & Social & Human & Other & Avg. \\
\midrule
Original &    21.0  &   23.6     &    28.2        &    21.3 & 23.5  \\ % 
Dense    &   28.7   &    35.5    &    30.1        &   31.6   & 31.5 \\ % 
SSD      &  30.8    &    34.1    &     32.7       &    30.3 & 32.0 \\ % 
\bottomrule
\end{tabular}
\vspace{-1em}
\end{table}

\subsection{Speed Analysis}

We report the average computation time of the original dense training and SMoE-based training in Table~\ref{tab:time}.
We use four NVIDIA A800 GPU for training and adopt MegaBlocks~\cite{DBLP:journals/corr/abs-2211-15841} as the SMoE implementation.
From this table, we observe that SSD achieve better time speedup on the larger model, i.e., Persimmon-8B.
This enhancement is attributed to the higher GPU utilization facilitated by the larger model, making the time speedup more obvious.
It highlights the promising speedup potential of SSD on LLMs.
However, a discrepancy is noted between theoretical speedup and actual time.
A deeper analysis into time consumption across different operations reveals that the expert selection process incurs additional time, thereby presenting an avenue for future research to optimize the computation of SMoE.
Although the time speedup of SSD is not fully matched with the theoretical speedup, the time reduction is indeed significant because the pre-training cost is huge and it would be increasingly valuable for large models.

\begin{table}[h]
\small
\centering
\caption{Speedup of SSD compared with dense training on the computational costs and times (hours).}
\label{tab:time}
\begin{tabular}{lcccc}
\toprule
& \multicolumn{2}{c}{GPT\quad\quad} & \multicolumn{2}{c}{Persimmon-8B\quad\quad} \\
        & Dense & SSD & Dense & SSD \\
\midrule
FLOPs &  237T       &  178T (1.44$\times$) &  2.97P       &  2.25P (1.32$\times$)   \\
\cmidrule{2-5}
Time  &   23.6h      &  21.5h (1.10$\times$)  &   166h       &  134h (1.23$\times$)   \\
\bottomrule
\end{tabular}
\end{table}

Here we also report the training time of other methods on GPT for comparison.
PLD takes 20.3 hours, MoE-Dropout takes 35.9 hours, and SSD takes 21.5 hours.
It show that our approach achieves competitive speedups compared with other methods.
It worth noting that the time speedups of acceleration methods can be affected by their specific implementations, making it challenging to directly compare their time speedups, so that we mainly focus on the FLOP speedups in the main text.

\subsection{Ablation Study}

In this subsection, we conduct ablation studies on the transition time determination and transition method. 
(1)~For the transition time determination, we compare the threshold-based method with the random method.
In the random method, we will switch the computation mode at each monitoring step with a probability of $0.5$.
We conduct this experiment on GPT three times and report the perplexity in Table~\ref{tab:ablation}.
From this table, we observe that the random method is consistently worse than the threshold-based method, which demonstrates the effectiveness of the threshold-based method.
The reason is that the random method may switch at the beginning of pre-training when the activation pattern is unstable, or switch frequently in a certain period, leading to unstable training.
(2)~For the transition method, we choose one switch point of GPT pre-training as the original model and apply different transition methods to it, including random, clustering, and SSD.
In the random method, we randomly split the neurons into groups.
In the clustering method, we directly use the clustering method in Section~\ref{sec:transition} without the previous checkpoint initialization as MoEfication does.
From Table~\ref{tab:ablation}, we observe that SSD achieves the smallest gap compared with the original model, which demonstrates the effectiveness of checkpoint initialization in SSD.
This minimal gap facilitates a smoother transition process, potentially culminating in superior performance.

\begin{table}[]
\small
\centering
\caption{Impact of transition time determination and transition method on the model perplexity.}
\label{tab:ablation}
\begin{tabular}{lcccc}
\toprule
\multirow{2.5}{*}{PPL} & \multicolumn{3}{c}{Random} & SSD  \\
\cmidrule(lr){2-4} \cmidrule(l){5-5}
 & 17.3    & 18.6    & 19.8   & \textbf{16.0} \\
\bottomrule
\toprule
\multirow{2.5}{*}{PPL} & \multicolumn{1}{c}{Original} & \multicolumn{1}{c}{Random} & Clustering & SSD  \\
\cmidrule(l){2-5}
 & \multicolumn{1}{c}{18.4}    & \multicolumn{1}{c}{3533.3}    & 20.9   & \textbf{20.4} \\
\bottomrule
\end{tabular}
\vspace{-1em}
\end{table}

\section{Conclusion}

In this work, we utilize the phenomenon of sparse activation to accelerate pre-training and inference of LLMs.
Specifically, we propose Switchable Sparse-Dense Learning, which adaptively switches between sparse and dense training.
Experimental results on three different model architectures and two kinds of downstream tasks show that our method can achieve comparable performance to dense training with less computational costs.
Moreover, the models trained with SSD can be directly used as MoE models for inference and reduce the inference time of FFNs by up to $2\times$ while keeping the performance as good as dense models.
We hope that our work can provide a new perspective for the acceleration based on sparse activation and inspire more research in this direction.

\section*{Impact Statement}

This paper presents work whose goal is to advance the field of Machine Learning. There are many potential societal consequences of our work, none of which we feel must be specifically highlighted here.

\section*{Acknowledgement}

This work is supported by the National Key R\&D
Program of China (No. 2022ZD0116312), Major Project of the National Social Science Foundation of China (No. 228ZD298), Quan Cheng Laboratory (Grant No. QCLZD202301), National Natural Science Foundation of China (No. 623B1019), Institute Guo Qiang at Tsinghua University and Beijing Advanced Innovation Center for Future Blockchain and Privacy Computing.

The method is primarily designed by Zhengyan Zhang, and he conducted most of the experiments. Chaojun Xiao, Qiujieli Qin, Yankai Lin, Zhiyuan Zeng, Xu Han, Zhiyuan Liu and Ruobing Xie actively participated in experimental design, result analysis, and paper writing. Maosong Sun led the project and provided guidance. Jie Zhou provided valuable advice to this research.

\bibliography{example_paper}

\begin{thebibliography}{69}
\providecommand{\natexlab}[1]{#1}
\providecommand{\url}[1]{\texttt{#1}}
\expandafter\ifx\csname urlstyle\endcsname\relax
  \providecommand{\doi}[1]{doi: #1}\else
  \providecommand{\doi}{doi: \begingroup \urlstyle{rm}\Url}\fi

\bibitem[Albericio et~al.(2016)Albericio, Judd, Hetherington, Aamodt, Jerger, and Moshovos]{CNNAct}
Albericio, J., Judd, P., Hetherington, T.~H., Aamodt, T.~M., Jerger, N. D.~E., and Moshovos, A.
\newblock Cnvlutin: Ineffectual-neuron-free deep neural network computing.
\newblock In \emph{Proceedings of ISCA}, pp.\  1--13, 2016.

\bibitem[Alizadeh et~al.(2023)Alizadeh, Mirzadeh, Belenko, Khatamifard, Cho, Mundo, Rastegari, and Farajtabar]{flash}
Alizadeh, K., Mirzadeh, I., Belenko, D., Khatamifard, K., Cho, M., Mundo, C. C.~D., Rastegari, M., and Farajtabar, M.
\newblock {LLM} in a flash: Efficient large language model inference with limited memory.
\newblock \emph{arXiv preprint arXiv:2312.11514}, 2023.

\bibitem[Artetxe et~al.(2022)Artetxe, Bhosale, Goyal, Mihaylov, Ott, Shleifer, Lin, Du, Iyer, Pasunuru, Anantharaman, Li, Chen, Akin, Baines, Martin, Zhou, Koura, O{'}Horo, Wang, Zettlemoyer, Diab, Kozareva, and Stoyanov]{artetxe-etal-2022-efficient}
Artetxe, M., Bhosale, S., Goyal, N., Mihaylov, T., Ott, M., Shleifer, S., Lin, X.~V., Du, J., Iyer, S., Pasunuru, R., Anantharaman, G., Li, X., Chen, S., Akin, H., Baines, M., Martin, L., Zhou, X., Koura, P.~S., O{'}Horo, B., Wang, J., Zettlemoyer, L., Diab, M., Kozareva, Z., and Stoyanov, V.
\newblock Efficient large scale language modeling with mixtures of experts.
\newblock In \emph{Proceedings of EMNLP}, pp.\  11699--11732, 2022.

\bibitem[Bowman et~al.(2015)Bowman, Angeli, Potts, and Manning]{snli}
Bowman, S.~R., Angeli, G., Potts, C., and Manning, C.~D.
\newblock A large annotated corpus for learning natural language inference.
\newblock In \emph{Proceedings of EMNLP}, pp.\  632--642, 2015.

\bibitem[Brown et~al.(2021)Brown, Mann, Ryder, Subbiah, Kaplan, Dhariwal, Neelakantan, Shyam, Sastry, Askell, Agarwal, Herbert-Voss, Krueger, Henighan, Child, Ramesh, Ziegler, Wu, Winter, Hesse, Chen, Sigler, Litwin, Gray, Chess, Clark, Berner, McCandlish, Radford, Sutskever, and Amodei]{GPT-3}
Brown, T.~B., Mann, B., Ryder, N., Subbiah, M., Kaplan, J., Dhariwal, P., Neelakantan, A., Shyam, P., Sastry, G., Askell, A., Agarwal, S., Herbert-Voss, A., Krueger, G., Henighan, T., Child, R., Ramesh, A., Ziegler, D.~M., Wu, J., Winter, C., Hesse, C., Chen, M., Sigler, E., Litwin, M., Gray, S., Chess, B., Clark, J., Berner, C., McCandlish, S., Radford, A., Sutskever, I., and Amodei, D.
\newblock Language models are {Few-Shot} learners.
\newblock In \emph{Proceedings of {NeurIPS}}, pp.\  1877--1901, 2021.

\bibitem[Chen et~al.(2022)Chen, Yin, Shang, Jiang, Qin, Wang, Wang, Chen, Liu, and Liu]{DBLP:conf/acl/ChenYS0QWWCL022}
Chen, C., Yin, Y., Shang, L., Jiang, X., Qin, Y., Wang, F., Wang, Z., Chen, X., Liu, Z., and Liu, Q.
\newblock bert2bert: Towards reusable pretrained language models.
\newblock In \emph{Proceedings of ACL}, pp.\  2134--2148, 2022.

\bibitem[Chen et~al.(2023{\natexlab{a}})Chen, Zhang, JAISWAL, Liu, and Wang]{moe-dropout}
Chen, T., Zhang, Z., JAISWAL, A.~K., Liu, S., and Wang, Z.
\newblock Sparse moe as the new dropout: Scaling dense and self-slimmable transformers.
\newblock In \emph{Proceedings of ICLR}, 2023{\natexlab{a}}.

\bibitem[Chen et~al.(2023{\natexlab{b}})Chen, Zhou, Du, Huang, Laudon, Chen, and Cui]{DBLP:conf/icml/ChenZDHLCC23}
Chen, W., Zhou, Y., Du, N., Huang, Y., Laudon, J., Chen, Z., and Cui, C.
\newblock Lifelong language pretraining with distribution-specialized experts.
\newblock In \emph{Proceedings of ICML}, 2023{\natexlab{b}}.

\bibitem[Chi et~al.(2022)Chi, Dong, Huang, Dai, Ma, Patra, Singhal, Bajaj, Song, Mao, Huang, and Wei]{DBLP:conf/nips/Chi0HDMPSBSMHW22}
Chi, Z., Dong, L., Huang, S., Dai, D., Ma, S., Patra, B., Singhal, S., Bajaj, P., Song, X., Mao, X., Huang, H., and Wei, F.
\newblock On the representation collapse of sparse mixture of experts.
\newblock In \emph{Proceedings of NeurIPS}, 2022.

\bibitem[Clark et~al.(2020)Clark, Luong, Le, and Manning]{electra}
Clark, K., Luong, M., Le, Q.~V., and Manning, C.~D.
\newblock {ELECTRA:} pre-training text encoders as discriminators rather than generators.
\newblock In \emph{Proceedings of ICLR}, 2020.

\bibitem[Devlin et~al.(2019)Devlin, Chang, Lee, and Toutanova]{BERT}
Devlin, J., Chang, M.-W., Lee, K., and Toutanova, K.
\newblock {{BERT}}: Pre-training of deep bidirectional transformers for language understanding.
\newblock In \emph{Proceedings of {NAACL}}, pp.\  4171--4186, 2019.

\bibitem[Dong et~al.(2023)Dong, Chen, and Chi]{dong2023towards}
Dong, H., Chen, B., and Chi, Y.
\newblock Towards structured sparsity in transformers for efficient inference.
\newblock In \emph{Workshop on Efficient Systems for Foundation Models @ ICML2023}, 2023.

\bibitem[Elsen et~al.(2023)Elsen, Odena, Nye, Ta\c{s}\i{}rlar, Dao, Hawthorne, Moparthi, and Somani]{persimmon-8b}
Elsen, E., Odena, A., Nye, M., Ta\c{s}\i{}rlar, S., Dao, T., Hawthorne, C., Moparthi, D., and Somani, A.
\newblock Releasing {Persimmon-8B}, 2023.

\bibitem[Fedus et~al.(2022)Fedus, Zoph, and Shazeer]{Switch-Transformer}
Fedus, W., Zoph, B., and Shazeer, N.
\newblock Switch transformers: Scaling to trillion parameter models with simple and efficient sparsity.
\newblock \emph{Journal of Machine Learning Research}, 23\penalty0 (120):\penalty0 1--39, 2022.

\bibitem[Gale et~al.(2022)Gale, Narayanan, Young, and Zaharia]{DBLP:journals/corr/abs-2211-15841}
Gale, T., Narayanan, D., Young, C., and Zaharia, M.
\newblock Megablocks: Efficient sparse training with mixture-of-experts.
\newblock In \emph{Proceedings of ICML}, 2022.

\bibitem[Gao et~al.(2021{\natexlab{a}})Gao, Biderman, Black, Golding, Hoppe, Foster, Phang, He, Thite, Nabeshima, Presser, and Leahy]{pile}
Gao, L., Biderman, S., Black, S., Golding, L., Hoppe, T., Foster, C., Phang, J., He, H., Thite, A., Nabeshima, N., Presser, S., and Leahy, C.
\newblock The pile: An 800gb dataset of diverse text for language modeling.
\newblock \emph{arxiv preprint arXiv:2101.00027}, 2021{\natexlab{a}}.

\bibitem[Gao et~al.(2021{\natexlab{b}})Gao, Tow, Biderman, Black, DiPofi, Foster, Golding, Hsu, McDonell, Muennighoff, Phang, Reynolds, Tang, Thite, Wang, Wang, and Zou]{eval-harness}
Gao, L., Tow, J., Biderman, S., Black, S., DiPofi, A., Foster, C., Golding, L., Hsu, J., McDonell, K., Muennighoff, N., Phang, J., Reynolds, L., Tang, E., Thite, A., Wang, B., Wang, K., and Zou, A.
\newblock A framework for few-shot language model evaluation, September 2021{\natexlab{b}}.

\bibitem[Gao et~al.(2022)Gao, Liu, Zhao, Lu, and Wen]{gao-etal-2022-parameter}
Gao, Z.-F., Liu, P., Zhao, W.~X., Lu, Z.-Y., and Wen, J.-R.
\newblock Parameter-efficient mixture-of-experts architecture for pre-trained language models.
\newblock In \emph{Proceedings of COLING}, pp.\  3263--3273, 2022.

\bibitem[Gong et~al.(2019)Gong, He, Li, Qin, Wang, and Liu]{DBLP:conf/icml/GongHLQWL19}
Gong, L., He, D., Li, Z., Qin, T., Wang, L., and Liu, T.
\newblock Efficient training of {BERT} by progressively stacking.
\newblock In \emph{Proceedings of ICML}, pp.\  2337--2346, 2019.

\bibitem[Gordon et~al.(2020)Gordon, Duh, and Andrews]{gordon-etal-2020-compressing}
Gordon, M., Duh, K., and Andrews, N.
\newblock Compressing {BERT}: Studying the effects of weight pruning on transfer learning.
\newblock In \emph{Proceedings of RL4NLP}, pp.\  143--155, 2020.

\bibitem[Gururangan et~al.(2022)Gururangan, Lewis, Holtzman, Smith, and Zettlemoyer]{gururangan-etal-2022-demix}
Gururangan, S., Lewis, M., Holtzman, A., Smith, N.~A., and Zettlemoyer, L.
\newblock {DEM}ix layers: Disentangling domains for modular language modeling.
\newblock In \emph{Proceedings of NAACL-HLT}, pp.\  5557--5576, 2022.

\bibitem[Hazimeh et~al.(2021)Hazimeh, Zhao, Chowdhery, Sathiamoorthy, Chen, Mazumder, Hong, and Chi]{DBLP:conf/nips/HazimehZCSCMHC21}
Hazimeh, H., Zhao, Z., Chowdhery, A., Sathiamoorthy, M., Chen, Y., Mazumder, R., Hong, L., and Chi, E.~H.
\newblock Dselect-k: Differentiable selection in the mixture of experts with applications to multi-task learning.
\newblock In \emph{Proceedings of NeurIPS}, pp.\  29335--29347, 2021.

\bibitem[Hendrycks \& Gimpel(2016)Hendrycks and Gimpel]{GeLU}
Hendrycks, D. and Gimpel, K.
\newblock Gaussian error linear units ({GELUs}).
\newblock \emph{arXiv preprint 1606.08415}, 2016.

\bibitem[Honovich et~al.(2022)Honovich, Scialom, Levy, and Schick]{UnnaturalInstructions}
Honovich, O., Scialom, T., Levy, O., and Schick, T.
\newblock Unnatural instructions: Tuning language models with (almost) no human labor.
\newblock \emph{arxiv preprint arXiv:2212.09689}, 2022.

\bibitem[Huang et~al.(2023)Huang, Bai, Zhu, Zhang, Zhang, Su, Liu, Lv, Zhang, Lei, Fu, Sun, and He]{DBLP:journals/corr/abs-2305-08322}
Huang, Y., Bai, Y., Zhu, Z., Zhang, J., Zhang, J., Su, T., Liu, J., Lv, C., Zhang, Y., Lei, J., Fu, Y., Sun, M., and He, J.
\newblock C-eval: {A} multi-level multi-discipline chinese evaluation suite for foundation models.
\newblock \emph{arXiv preprint arXiv:2305.08322}, 2023.

\bibitem[Hwang et~al.(2022)Hwang, Cui, Xiong, Yang, Liu, Hu, Wang, Salas, Jose, Ram, Chau, Cheng, Yang, Yang, and Xiong]{DBLP:journals/corr/abs-2206-03382}
Hwang, C., Cui, W., Xiong, Y., Yang, Z., Liu, Z., Hu, H., Wang, Z., Salas, R., Jose, J., Ram, P., Chau, J., Cheng, P., Yang, F., Yang, M., and Xiong, Y.
\newblock Tutel: Adaptive mixture-of-experts at scale.
\newblock \emph{arXiv preprint arXiv:2206.03382}, 2022.

\bibitem[Izsak et~al.(2021)Izsak, Berchansky, and Levy]{DBLP:conf/emnlp/IzsakBL21}
Izsak, P., Berchansky, M., and Levy, O.
\newblock How to train {BERT} with an academic budget.
\newblock In \emph{Proceedings of EMNLP}, pp.\  10644--10652, 2021.

\bibitem[Jang et~al.(2023)Jang, Kim, Ye, Kim, Logeswaran, Lee, Lee, and Seo]{DBLP:conf/icml/JangKYKLLLS23}
Jang, J., Kim, S., Ye, S., Kim, D., Logeswaran, L., Lee, M., Lee, K., and Seo, M.
\newblock Exploring the benefits of training expert language models over instruction tuning.
\newblock In \emph{Proceedings of ICML}, 2023.

\bibitem[Jiang et~al.(2024)Jiang, Sablayrolles, Roux, Mensch, Savary, Bamford, Chaplot, de~Las~Casas, Hanna, Bressand, Lengyel, Bour, Lample, Lavaud, Saulnier, Lachaux, Stock, Subramanian, Yang, Antoniak, Scao, Gervet, Lavril, Wang, Lacroix, and Sayed]{mixtral}
Jiang, A.~Q., Sablayrolles, A., Roux, A., Mensch, A., Savary, B., Bamford, C., Chaplot, D.~S., de~Las~Casas, D., Hanna, E.~B., Bressand, F., Lengyel, G., Bour, G., Lample, G., Lavaud, L.~R., Saulnier, L., Lachaux, M., Stock, P., Subramanian, S., Yang, S., Antoniak, S., Scao, T.~L., Gervet, T., Lavril, T., Wang, T., Lacroix, T., and Sayed, W.~E.
\newblock Mixtral of experts.
\newblock \emph{arXiv preprint arXiv:2401.04088}, 2024.

\bibitem[Johnson et~al.(2019)Johnson, Douze, and J{\'e}gou]{johnson2019billion}
Johnson, J., Douze, M., and J{\'e}gou, H.
\newblock Billion-scale similarity search with {GPUs}.
\newblock \emph{IEEE Transactions on Big Data}, 7\penalty0 (3):\penalty0 535--547, 2019.

\bibitem[Kingma \& Ba(2015)Kingma and Ba]{adam}
Kingma, D.~P. and Ba, J.
\newblock Adam: {A} method for stochastic optimization.
\newblock In \emph{Proceedings of ICLR}, 2015.

\bibitem[Lee-Thorp \& Ainslie(2022)Lee-Thorp and Ainslie]{lee-thorp-ainslie-2022-sparse}
Lee-Thorp, J. and Ainslie, J.
\newblock Sparse mixers: Combining {M}o{E} and mixing to build a more efficient {BERT}.
\newblock In \emph{Findings of EMNLP}, pp.\  58--75, 2022.

\bibitem[Lewis et~al.(2021)Lewis, Bhosale, Dettmers, Goyal, and Zettlemoyer]{baselayers}
Lewis, M., Bhosale, S., Dettmers, T., Goyal, N., and Zettlemoyer, L.
\newblock {BASE} layers: Simplifying training of large, sparse models.
\newblock In \emph{Proceedings of ICM}, volume 139, pp.\  6265--6274, 2021.

\bibitem[Li et~al.(2023)Li, You, Bhojanapalli, Li, Rawat, Reddi, Ye, Chern, Yu, Guo, and Kumar]{li2023the}
Li, Z., You, C., Bhojanapalli, S., Li, D., Rawat, A.~S., Reddi, S.~J., Ye, K., Chern, F., Yu, F., Guo, R., and Kumar, S.
\newblock The lazy neuron phenomenon: On emergence of activation sparsity in transformers.
\newblock In \emph{Proceedings of ICLR}, 2023.

\bibitem[Lin(2004)]{rouge}
Lin, C.-Y.
\newblock {ROUGE}: A package for automatic evaluation of summaries.
\newblock In \emph{Text Summarization Branches Out}, pp.\  74--81, July 2004.

\bibitem[Liu et~al.(2022)Liu, Kim, Muzio, and Hassan]{DBLP:conf/icml/0013KMH22}
Liu, R., Kim, Y.~J., Muzio, A., and Hassan, H.
\newblock Gating dropout: Communication-efficient regularization for sparsely activated transformers.
\newblock In \emph{International Conference on Machine Learning, {ICML} 2022, 17-23 July 2022, Baltimore, Maryland, {USA}}, Proceedings of ICML, 2022.

\bibitem[Liu et~al.(2023)Liu, Wang, Dao, Zhou, Yuan, Song, Shrivastava, Zhang, Tian, R{\'{e}}, and Chen]{dejavu}
Liu, Z., Wang, J., Dao, T., Zhou, T., Yuan, B., Song, Z., Shrivastava, A., Zhang, C., Tian, Y., R{\'{e}}, C., and Chen, B.
\newblock Deja vu: Contextual sparsity for efficient llms at inference time.
\newblock In Krause, A., Brunskill, E., Cho, K., Engelhardt, B., Sabato, S., and Scarlett, J. (eds.), \emph{Preceedings of ICML}, volume 202, pp.\  22137--22176, 2023.

\bibitem[Malinen \& Fr{\"{a}}nti(2014)Malinen and Fr{\"{a}}nti]{bkmeans}
Malinen, M.~I. and Fr{\"{a}}nti, P.
\newblock Balanced k-means for clustering.
\newblock In \emph{Proceedings of SSSPR}, volume 8621, pp.\  32--41, 2014.

\bibitem[Mirzadeh et~al.(2023)Mirzadeh, Alizadeh, Mehta, Mundo, Tuzel, Samei, Rastegari, and Farajtabar]{mirzadeh2023relu}
Mirzadeh, I., Alizadeh, K., Mehta, S., Mundo, C. C.~D., Tuzel, O., Samei, G., Rastegari, M., and Farajtabar, M.
\newblock Relu strikes back: Exploiting activation sparsity in large language models.
\newblock \emph{arXiv preprint arXiv:2310.04564}, 2023.

\bibitem[Muqeeth et~al.(2023)Muqeeth, Liu, and Raffel]{DBLP:journals/corr/abs-2306-03745}
Muqeeth, M., Liu, H., and Raffel, C.
\newblock Soft merging of experts with adaptive routing.
\newblock \emph{arXiv preprint arXiv:2306.03745}, 2023.

\bibitem[Pan et~al.(2024)Pan, Shen, Liu, Mishra, Zhang, Oliva, Raffel, and Panda]{DBLP:journals/corr/abs-2404-05567}
Pan, B., Shen, Y., Liu, H., Mishra, M., Zhang, G., Oliva, A., Raffel, C., and Panda, R.
\newblock Dense training, sparse inference: Rethinking training of mixture-of-experts language models.
\newblock \emph{arXiv preprint arXiv:2404.05567}, 2024.

\bibitem[Qin et~al.(2022)Qin, Lin, Yi, Zhang, Han, Zhang, Su, Liu, Li, Sun, and Zhou]{DBLP:conf/naacl/QinLYZ0ZSLLS022}
Qin, Y., Lin, Y., Yi, J., Zhang, J., Han, X., Zhang, Z., Su, Y., Liu, Z., Li, P., Sun, M., and Zhou, J.
\newblock Knowledge inheritance for pre-trained language models.
\newblock In \emph{Proceedings of NAACL-HLT}, pp.\  3921--3937, 2022.

\bibitem[Radford et~al.(2019)Radford, Wu, Child, Luan, Amodei, and Sutskever]{GPT-2}
Radford, A., Wu, J., Child, R., Luan, D., Amodei, D., and Sutskever, I.
\newblock Language models are unsupervised multitask learners.
\newblock 2019.

\bibitem[Raffel et~al.(2020)Raffel, Shazeer, Roberts, Lee, Narang, Matena, Zhou, Li, and Liu]{T5}
Raffel, C., Shazeer, N., Roberts, A., Lee, K., Narang, S., Matena, M., Zhou, Y., Li, W., and Liu, P.~J.
\newblock Exploring the limits of transfer learning with a unified {Text-to-Text} transformer.
\newblock \emph{J. Mach. Learn. Res.}, 21:\penalty0 140:1--140:67, 2020.

\bibitem[Rajbhandari et~al.(2022)Rajbhandari, Li, Yao, Zhang, Aminabadi, Awan, Rasley, and He]{DBLP:conf/icml/RajbhandariLYZA22}
Rajbhandari, S., Li, C., Yao, Z., Zhang, M., Aminabadi, R.~Y., Awan, A.~A., Rasley, J., and He, Y.
\newblock Deepspeed-moe: Advancing mixture-of-experts inference and training to power next-generation {AI} scale.
\newblock In \emph{Proceedings of ICML}, 2022.

\bibitem[Rajpurkar et~al.(2016)Rajpurkar, Zhang, Lopyrev, and Liang]{rajpurkar2016squad}
Rajpurkar, P., Zhang, J., Lopyrev, K., and Liang, P.
\newblock {SQ}u{AD}: 100,000+ questions for machine comprehension of text.
\newblock In \emph{Proceedings of EMNLP}, pp.\  2383--2392. Association for Computational Linguistics, 2016.

\bibitem[Rand(1971)]{ARI}
Rand, W.~M.
\newblock Objective criteria for the evaluation of clustering methods.
\newblock \emph{Journal of the American Statistical association}, 66\penalty0 (336):\penalty0 846--850, 1971.

\bibitem[Riquelme et~al.(2021)Riquelme, Puigcerver, Mustafa, Neumann, Jenatton, Pinto, Keysers, and Houlsby]{DBLP:conf/nips/RiquelmePMNJPKH21}
Riquelme, C., Puigcerver, J., Mustafa, B., Neumann, M., Jenatton, R., Pinto, A.~S., Keysers, D., and Houlsby, N.
\newblock Scaling vision with sparse mixture of experts.
\newblock In \emph{Proceedings of NeurIPS}, 2021.

\bibitem[Roller et~al.(2021)Roller, Sukhbaatar, Szlam, and Weston]{HashLayers}
Roller, S., Sukhbaatar, S., Szlam, A., and Weston, J.
\newblock Hash layers for large sparse models.
\newblock In \emph{Proceedings of NeurIPS}, pp.\  17555--17566, 2021.

\bibitem[Seung(2000)]{seung2000half}
Seung, H.~S.
\newblock Half a century of hebb.
\newblock \emph{Nature neuroscience}, 3\penalty0 (11):\penalty0 1166--1166, 2000.

\bibitem[Shazeer(2020)]{GLU}
Shazeer, N.
\newblock {GLU} variants improve transformer.
\newblock \emph{arXiv preprint arXiv:2002.05202}, 2020.

\bibitem[Socher et~al.(2013)Socher, Perelygin, Wu, Chuang, Manning, Ng, and Potts]{socher2013recursive}
Socher, R., Perelygin, A., Wu, J., Chuang, J., Manning, C.~D., Ng, A., and Potts, C.
\newblock Recursive deep models for semantic compositionality over a sentiment treebank.
\newblock In \emph{Proceedings of EMNLP}, pp.\  1631--1642, 2013.

\bibitem[Tan et~al.(2024)Tan, Shen, Panda, and Courville]{ScatterMoE}
Tan, S., Shen, Y., Panda, R., and Courville, A.~C.
\newblock Scattered mixture-of-experts implementation.
\newblock \emph{arXiv preprint arXiv:2301.11514}, 2024.

\bibitem[Touvron et~al.(2023{\natexlab{a}})Touvron, Lavril, Izacard, Martinet, Lachaux, Lacroix, Rozi{\`{e}}re, Goyal, Hambro, Azhar, Rodriguez, Joulin, Grave, and Lample]{llama}
Touvron, H., Lavril, T., Izacard, G., Martinet, X., Lachaux, M., Lacroix, T., Rozi{\`{e}}re, B., Goyal, N., Hambro, E., Azhar, F., Rodriguez, A., Joulin, A., Grave, E., and Lample, G.
\newblock Llama: Open and efficient foundation language models.
\newblock \emph{arxiv preprint arXiv:2302.13971}, 2023{\natexlab{a}}.

\bibitem[Touvron et~al.(2023{\natexlab{b}})Touvron, Martin, Stone, Albert, et~al.]{LLAMA2}
Touvron, H., Martin, L., Stone, K., Albert, P., et~al.
\newblock Llama 2: Open foundation and fine-tuned chat models.
\newblock \emph{arxiv preprint arXiv:2307.09288}, 2023{\natexlab{b}}.

\bibitem[Vaswani et~al.(2017{\natexlab{a}})Vaswani, Shazeer, Parmar, and Uszkoreit]{Transformer}
Vaswani, A., Shazeer, N., Parmar, N., and Uszkoreit, J.
\newblock Attention is all you need.
\newblock In \emph{Proceedings of {NeurIPS}}, pp.\  5998--6008, 2017{\natexlab{a}}.

\bibitem[Vaswani et~al.(2017{\natexlab{b}})Vaswani, Shazeer, Parmar, Uszkoreit, Jones, Gomez, Kaiser, and Polosukhin]{transformer-orig}
Vaswani, A., Shazeer, N., Parmar, N., Uszkoreit, J., Jones, L., Gomez, A.~N., Kaiser, L., and Polosukhin, I.
\newblock Attention is all you need.
\newblock In \emph{Proceedings of NeurIPS}, pp.\  5998--6008, 2017{\natexlab{b}}.

\bibitem[Wang et~al.(2022{\natexlab{a}})Wang, Wen, Zhang, Hou, Liu, and Li]{SkillNeuron}
Wang, X., Wen, K., Zhang, Z., Hou, L., Liu, Z., and Li, J.
\newblock Finding skill neurons in pre-trained transformer-based language models.
\newblock In \emph{Proceedings of EMNLP}, 2022{\natexlab{a}}.

\bibitem[Wang et~al.(2022{\natexlab{b}})Wang, Mishra, Alipoormolabashi, Kordi, Mirzaei, Arunkumar, Ashok, Dhanasekaran, Naik, Stap, et~al.]{supernaturalinstructions}
Wang, Y., Mishra, S., Alipoormolabashi, P., Kordi, Y., Mirzaei, A., Arunkumar, A., Ashok, A., Dhanasekaran, A.~S., Naik, A., Stap, D., et~al.
\newblock Super-naturalinstructions:generalization via declarative instructions on 1600+ tasks.
\newblock In \emph{EMNLP}, 2022{\natexlab{b}}.

\bibitem[Williams et~al.(2018)Williams, Nangia, and Bowman]{mnli}
Williams, A., Nangia, N., and Bowman, S.~R.
\newblock A broad-coverage challenge corpus for sentence understanding through inference.
\newblock In \emph{Proceedings of NAACL-HLT}, 2018.

\bibitem[Xiong et~al.(2020)Xiong, Yang, He, Zheng, Zheng, Xing, Zhang, Lan, Wang, and Liu]{pre-ln}
Xiong, R., Yang, Y., He, D., Zheng, K., Zheng, S., Xing, C., Zhang, H., Lan, Y., Wang, L., and Liu, T.
\newblock On layer normalization in the transformer architecture.
\newblock In \emph{Proceedings of ICML}, pp.\  10524--10533, 2020.

\bibitem[Yang et~al.(2022)Yang, Zhang, Lan, Yang, Li, Tan, Xiao, and Pu]{DBLP:conf/mm/YangZLYLTXP22}
Yang, Q., Zhang, K., Lan, C., Yang, Z., Li, Z., Tan, W., Xiao, J., and Pu, S.
\newblock Unified normalization for accelerating and stabilizing transformers.
\newblock In \emph{Proceedings of MM}, pp.\  4445--4455, 2022.

\bibitem[Zhang \& He(2020)Zhang and He]{PLD}
Zhang, M. and He, Y.
\newblock Accelerating training of transformer-based language models with progressive layer dropping.
\newblock In \emph{Proceedings of NeurIPS}, 2020.

\bibitem[Zhang et~al.(2022{\natexlab{a}})Zhang, Shen, Huang, Zhou, Rong, and Xiong]{moa}
Zhang, X., Shen, Y., Huang, Z., Zhou, J., Rong, W., and Xiong, Z.
\newblock Mixture of attention heads: Selecting attention heads per token.
\newblock In \emph{Proceedings of EMNLP}, pp.\  4150--4162, 2022{\natexlab{a}}.

\bibitem[Zhang et~al.(2022{\natexlab{b}})Zhang, Lin, Liu, Li, Sun, and Zhou]{zhang2022moefication}
Zhang, Z., Lin, Y., Liu, Z., Li, P., Sun, M., and Zhou, J.
\newblock {MoEfication}: Transformer feed-forward layers are mixtures of experts.
\newblock In \emph{Findings of ACL}, 2022{\natexlab{b}}.

\bibitem[Zhang et~al.(2023)Zhang, Zeng, Lin, Xiao, Wang, Han, Liu, Xie, Sun, and Zhou]{zhang2023emergent}
Zhang, Z., Zeng, Z., Lin, Y., Xiao, C., Wang, X., Han, X., Liu, Z., Xie, R., Sun, M., and Zhou, J.
\newblock Emergent modularity in pre-trained transformers.
\newblock In \emph{Findings of ACL}, 2023.

\bibitem[Zhang et~al.(2024)Zhang, Song, Yu, Han, Lin, Xiao, Song, Liu, Mi, and Sun]{relu2}
Zhang, Z., Song, Y., Yu, G., Han, X., Lin, Y., Xiao, C., Song, C., Liu, Z., Mi, Z., and Sun, M.
\newblock Relu\({}^{\mbox{2}}\) wins: Discovering efficient activation functions for sparse llms.
\newblock \emph{arXiv preprint arXiv:2402.03804}, 2024.

\bibitem[Zuo et~al.(2022{\natexlab{a}})Zuo, Liu, Jiao, Kim, Hassan, Zhang, Gao, and Zhao]{THOR}
Zuo, S., Liu, X., Jiao, J., Kim, Y.~J., Hassan, H., Zhang, R., Gao, J., and Zhao, T.
\newblock Taming sparsely activated transformer with stochastic experts.
\newblock In \emph{Proceedings of ICLR}, 2022{\natexlab{a}}.

\bibitem[Zuo et~al.(2022{\natexlab{b}})Zuo, Zhang, Liang, He, Zhao, and Chen]{zuo-etal-2022-moebert}
Zuo, S., Zhang, Q., Liang, C., He, P., Zhao, T., and Chen, W.
\newblock {M}o{EBERT}: from {BERT} to mixture-of-experts via importance-guided adaptation.
\newblock In \emph{Proceedings of NAACL-HLT}, pp.\  1610--1623, 2022{\natexlab{b}}.

\end{thebibliography}
\bibliographystyle{icml2024}

\newpage
\appendix
\onecolumn

\section{Other Experimental Details}
\label{sec:appendix}

\textbf{Model Architecture.} We use layer normalization before each attention and FFN layer~\cite{pre-ln}, which is beneficial for convergence~\cite{DBLP:conf/emnlp/IzsakBL21}, and use ReLU as the activation function for easier MoE transformation than GeLU~\cite{GeLU,zhang2022moefication}.

\textbf{Pre-training.} 
We use part of the Pile dataset~\cite{pile} as the pre-training corpus.
Based on observations from previous work~\cite{zhang2023emergent}, the emergence of sparse activations appears to be more closely tied to the inherent training dynamics arising from the model architecture itself, rather than the type or scale of the data used. Therefore, we believe that using a subset of the Pile dataset for our pretraining experiments does not significantly impact the degree of activation sparsity observed.
We use the Adam optimizer~\cite{adam} and Noam learning rate scheduler~\cite{Transformer} for pre-training.
The batch size is set to 512 and the learning rate is set to 1 for BERT and T5 and 0.5 for GPT.
The mask rate of MLM is set to 0.15.
PLD increases the overall dropout rate from 0 to 0.25 quickly at the beginning of training and then keeps it at 0.25.
For the MoE layers, we set the number of experts $N$ to 32 for MoE-Dropout and SSD.
MoE-Dropout linearly increases the number of selected experts $K$ from 6 to 32 during the pre-training.
For SSD, we set the threshold $\tau$ to 0.9 and monitor the activation pattern every 3,000 steps. In the sparse mode, we also select 6 experts for each layer. The ratio of the sparse mode $r$ is set to 0.5. The ratio of the final dense training $l$ is set to 0.1.
For SMoE, we set the number of experts $N$ to 3 and the number of selected experts $K$ to 2 to ensure the computational cost is similar to that of other methods.

\textbf{Persimmon-8B.} For the continual pre-training of Persimmon-8B, we set the batch size to 2,048, the learning rate to 0.00003, and the max sequence length to 1024.
The total pre-training steps are set to 128,000 and the gradient accumulation steps are set to 64.
As a result, the total number of optimization steps is 2000.
We use the same ratio of the sparse mode $r=0.5$ as the experiments on base-scale models.
Since the total optimization steps are smaller than training from scratch, we divide the SSD training into two stages: the first stage is 1000 steps of sparse training and the second stage is 1000 steps of dense training.
We compare the SSD performance on Persimmon-8B with the dense training performance on Persimmon-8B with the same optimization steps.
The total expert number $N$ is set to 64 and the selected expert number $K$ is set to 16, keeping the expert size the same as that of the base models.
The other hyperparameters are the same as those of base models.
When evaluating the performance of Persimmon-8B, we use LM Evaluation Harness~\cite{eval-harness}.

\textbf{Downstream Tasks.}
First, we evaluate models on several natural language understanding tasks.
For single-sentence classification, we use SST-2~\cite{socher2013recursive}, which is a sentiment analysis dataset.
For sentence-pair classification, we use SNLI~\cite{snli}, MNLI~\cite{mnli}, QNLI~\cite{rajpurkar2016squad}, and QQP\footnote{\url{https://data.quora.com/First-Quora-Dataset-Release-Question-Pairs}}, covering the tasks of natural language inference and paraphrase identification.
For reading comprehension, we use SQuAD v1.1~\cite{rajpurkar2016squad}, which is a widely used dataset for extractive question answering.
Second, we evaluate models on instruction tuning. 
Specifically, we follow the setups of \citet{UnnaturalInstructions}, where the model is trained on a model-generated instruction dataset and evaluated on several human-labeled instruction datasets.
The training dataset is Unnatural Instructions~\cite{UnnaturalInstructions} and the development dataset contains 1,000 randomly sampled instances from the training set of Super-NaturalInstructions~\cite{supernaturalinstructions}, which is used to select the best checkpoint.
The test dataset is the test set of Super-NaturalInstructions~\cite{supernaturalinstructions}. 
We conduct a grid search to find the best hyperparameters for each model, including the learning rate varied from 4e-4 to 2e-3, the batch size varied from 16 to 32, and the number of training epochs varied from 3 to 10.

\textbf{SMoE Implementation.} 
There are some frameworks optimized for SMoE, such as DeepSpeed-MoE~\cite{DBLP:conf/icml/RajbhandariLYZA22}, MegaBlocks~\cite{DBLP:journals/corr/abs-2211-15841}, and Tutel~\cite{DBLP:journals/corr/abs-2206-03382}.
Among them, we implement SMoE based on the MegaBlocks framework, which supports efficient dropless MoE layers.

\textbf{Evaluation of Instruction Tuning.}
Following~\citet{UnnaturalInstructions}, we adopt greedy decoding to generate the responses to the instructions.
Super-NaturalInstructions~\cite{supernaturalinstructions} is evaluated by Rouge-L~\cite{rouge}.

\textbf{WCSS Computation.}  In the paper, we follow the standard formulation for calculating WCSS: $\sum_{k=1}^{K} \sum_{i \in S_{k}} \sum_{j=1}^{p}(x_{i j}-\bar{x}_{k j})^{2}$, where $S_k$ is the set of data points assigned to cluster $k$, $x_{i j}$ is the $j$-th feature value of data point $i$, and $\bar{x}_{k j}$ is the mean value of feature $j$ across all points in cluster $k$.
In our case of WCSS clustering, $x_i$ is the $i$-th row of $\mW_i$.

\textbf{Post-processing of Expert Scores.} In the paper, we set the $\alpha_n$ of selected experts to $1$ through post-processing.
Specifically, we compute $\alpha = 1 + \alpha - \alpha\text{.detach()}$ in Pytorch to ensure the gradient backpropagation.

\textbf{Clustering.} The clustering operation is performed on the GPU using the faiss-gpu library~\cite{johnson2019billion}.
To ensure the balance of the number of neurons in each expert, we further apply the balanced assignment strategy~\cite{baselayers} to the results of k-means clustering.
The time of each clustering is less than 1 minute and has little impact on the training time.

\begin{table}[h]
\small
\centering
\caption{Evaluation results (\%) on natural language understanding tasks. MoE-D refers to MoE-Dropout. SMoE (D) refers to densely fine-tuned SMoE. Since the training costs of SMoE are smaller than other methods, we color the results of SMoE in {\color{gray} gray}. The last column represents the average of all previous columns.}
\label{tab:nlu-std}
\begin{tabular}{l|rrrrrrr|r}
\toprule
& \multicolumn{1}{c}{SST2} & \multicolumn{1}{c}{SNLI} & \multicolumn{1}{c}{MNLI-m} & \multicolumn{1}{c}{QNLI} & \multicolumn{1}{c}{QQP} & \multicolumn{2}{c|}{SQuAD} & \multirow{2}{*}{Avg.} \\
& \multicolumn{1}{c}{Acc.}  & \multicolumn{1}{c}{Acc.}  & \multicolumn{1}{c}{Acc.}  & \multicolumn{1}{c}{Acc}  & \multicolumn{1}{c}{Acc.} & \multicolumn{1}{c}{EM}          & \multicolumn{1}{c|}{F1}    &      \\
\midrule
\multicolumn{9}{c}{BERT-based Models} \\ \midrule
Dense &  90.0$\pm0.5$    &   88.4$\pm0.1$   &     80.1$\pm0.1$ &   87.2$\pm0.4$   &  89.7$\pm0.1$   &      68.4$\pm0.2$       &    78.9$\pm0.2$  & 83.2       \\ % 
{\color{gray} SMoE} & {\color{gray} 84.3$\pm0.2$} & {\color{gray} 83.9$\pm0.2$} & {\color{gray} 71.4$\pm0.7$} & {\color{gray} 82.3$\pm0.2$} & {\color{gray} 85.9$\pm0.7$} & {\color{gray} 56.8$\pm1.0$} & {\color{gray} 68.8$\pm0.7$} & {\color{gray} 76.2} \\ % 
SMoE (D) & 84.7$\pm1.0$ & 84.5$\pm0.1$ & 71.8$\pm0.2$ & 82.8$\pm0.1$ & 86.5$\pm0.4$ & 59.0$\pm1.0$ & 70.9$\pm0.8$ & 77.2 \\ % 
PLD   &   90.0$\pm0.4$   &  88.3$\pm0.2$    &  79.5$\pm0.3$    &   86.6$\pm0.4$   &  89.8$\pm0.1$   &     67.3$\pm0.3$         &     77.9$\pm0.3$  & 82.8      \\ % 
MoE-D &  90.7$\pm0.3$    &   88.5$\pm0.2$   & 80.7$\pm0.2$     &  87.4$\pm0.3$    &  89.8$\pm0.1$   &     65.9$\pm0.4$        &       76.8$\pm0.7$  & 82.8    \\ % 
SSD   & 90.6$\pm0.4$    &  88.7$\pm0.1$    &  80.6$\pm0.4$    &  88.7$\pm0.4$    &  90.0$\pm0.1$   &    68.1$\pm0.3$         &     78.8$\pm0.1$  & 83.6      \\ \midrule % 
\multicolumn{9}{c}{T5-based Models} \\ \midrule
Dense &   91.5$\pm0.5$   &   89.4$\pm0.2$   &   81.7$\pm0.1$   &  88.8$\pm0.2$    &   90.2$\pm0.1$  &   75.5$\pm0.4$          &  83.9$\pm0.3$   & 85.9        \\ % 
{\color{gray} SMoE} & {\color{gray} 86.0$\pm0.1$} & {\color{gray} 87.5$\pm0.1$} & {\color{gray} 77.9$\pm0.4$} & {\color{gray} 84.2$\pm0.0$} & {\color{gray} 88.3$\pm0.2$} & {\color{gray} 68.8$\pm0.3$} & {\color{gray} 78.5$\pm0.3$} & {\color{gray} 81.6}\\ % 
SMoE (D) & 86.2$\pm0.6$ & 87.8$\pm0.2$ & 78.6$\pm0.5$ & 84.7$\pm0.2$ & 88.6$\pm0.1$ & 69.1$\pm0.3$ & 78.8$\pm0.2$ & 82.0\\ % 
PLD   &   92.1$\pm0.1$   &  89.5$\pm0.1$    &   82.8$\pm0.1$   &  89.9$\pm0.1$    &  90.4$\pm0.1$   &  76.8$\pm0.3$           &     85.1$\pm0.4$   & 86.7     \\ % 
MoE-D &   91.8$\pm0.3$    &   89.4$\pm0.1$   &  82.3$\pm0.1$    &   89.5$\pm0.2$   &   90.4$\pm0.1$  &   74.9$\pm0.2$          &   83.6$\pm0.1$   & 86.0  \\ % 
SSD   &   92.5$\pm0.2$    &  89.8$\pm0.1$    &   82.5$\pm0.1$   &  89.5$\pm0.1$    &   90.5$\pm0.1$  &         76.5$\pm0.4$    &        85.0$\pm0.2$   & 86.6 \\ % 
\bottomrule
\end{tabular}
\end{table}

\section{Results with Standard Deviation}
\label{sec:std}

We report the evaluation results with standard deviation on natural language understanding tasks in Table~\ref{tab:nlu-std} and instruction tuning tasks in Table~\ref{tab:instruction-std}.

\begin{table}[h]
\small
\centering
\caption{Evaluation results (\%) on instruction tuning. MoE-D refers to MoE-Dropout. SMoE (D) refers to densely fine-tuned SMoE. ``Dev'' represents the development set and ``NI'' represents Super-NaturalInstructions. Since the training costs of SMoE are smaller than other methods, we color the results of SMoE in {\color{gray} gray}. The last column represents the average result.}
\label{tab:instruction-std}
\begin{tabular}{lrrr}
\toprule
& \multicolumn{1}{c}{Dev}     & \multicolumn{1}{c}{NI}       & \multirow{2}{*}{Avg.}  \\
& \multicolumn{1}{c}{Rouge-L} & \multicolumn{1}{c}{Rouge-L}  & \\
\midrule
\multicolumn{4}{c}{GPT-based Models} \\ \midrule
Dense &    19.2$\pm0.3$     &   16.7$\pm0.6$      &  18.0 \\ % 
{\color{gray} SMoE} & {\color{gray} 17.6$\pm0.4$} & {\color{gray} 15.9$\pm0.8$} & {\color{gray} 16.8} \\ % 
SMoE (D) & 18.3$\pm0.5$ & 17.9$\pm0.9$ & 18.1 \\ % 
PLD   &  19.1$\pm0.7$       &  17.5$\pm0.4$       &  18.3 \\ % 
MoE-D &   18.8$\pm0.5$      &   18.2$\pm1.7$      &   18.5 \\ % 
SSD   &  19.4$\pm0.4$       &    18.0$\pm0.9$    &   18.7 \\ \midrule % 
\multicolumn{4}{c}{T5-based Models} \\ \midrule
Dense &      19.7$\pm0.3$   &    19.1$\pm0.8$     &  19.4 \\ % 
{\color{gray} SMoE} & {\color{gray} 16.7$\pm0.2$} & {\color{gray} 15.8$\pm0.2$} & {\color{gray} 16.3} \\ % 
SMoE (D) & 17.0$\pm0.7$ & 16.3$\pm1.0$ & 16.7 \\ % 
PLD   &     19.5$\pm0.3$   &    20.1$\pm0.7$     &  19.8 \\ % 
MoE-D &      18.6$\pm0.3$   &    18.9$\pm0.7$     &  18.7 \\ % 
SSD   &      18.6$\pm0.2$   &    20.4$\pm0.6$     &  19.5 \\ \bottomrule % 
\end{tabular}
\end{table}

\section{Case Study}

\begin{figure}[h]
\centering
\includegraphics[width=0.5\linewidth]{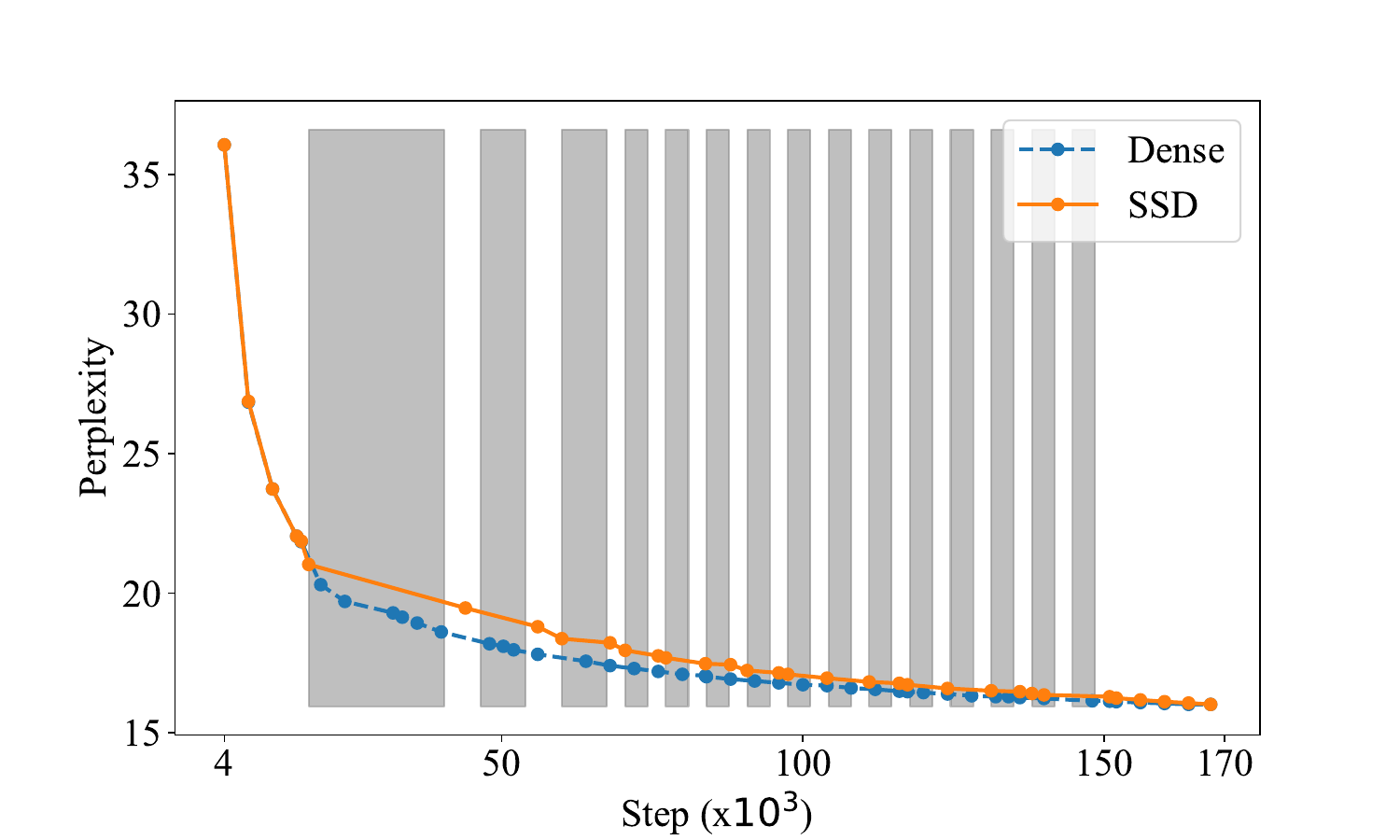}
\caption{Change of evaluation perplexity during pre-training. The model architecture is GPT. The gray areas represent the stages of sparse mode for SSD.}
\label{fig:case-loss}
\end{figure}

To better understand the training process of SSD, we visualize the change of evaluation perplexity and the stages of sparse mode in Figure~\ref{fig:case-loss}.
From this figure, we observe that there are three different lengths of sparse stages: the first one is the longest~(22,500 steps), the second and third ones are the second longest~(7,500 steps), and the rest are the shortest~(3,750 steps).
Here the proportion of the sparse mode to the dense mode is set to 1.25.
The first dense training has 18,000 steps, followed by 22,500 ($18000\times1.25$) sparse training steps. 
The second dense training has 6,000 steps, followed by 7,500 ($6000\times1.25$) sparse training steps.
It reveals the change of activation pattern during pre-training, i.e., the activation pattern changes dramatically at the beginning and then becomes stable.
When the activation pattern is unstable, the length of the dense stage is long, and the corresponding sparse stage is also long, vice versa.
Mixing dense training with sparse training will slow down the learning of the model, and the perplexity is higher than that of dense training in the middle stage.
However, the perplexity of SSD can quickly catch up with that of dense training after the final dense training.
Note that at the same step, the computational cost of SSD is much lower than that of dense training, which indicates that SSD can achieve a good trade-off between computational cost and model performance.

\begin{figure}[h]
\centering
\includegraphics[width=0.5\linewidth]{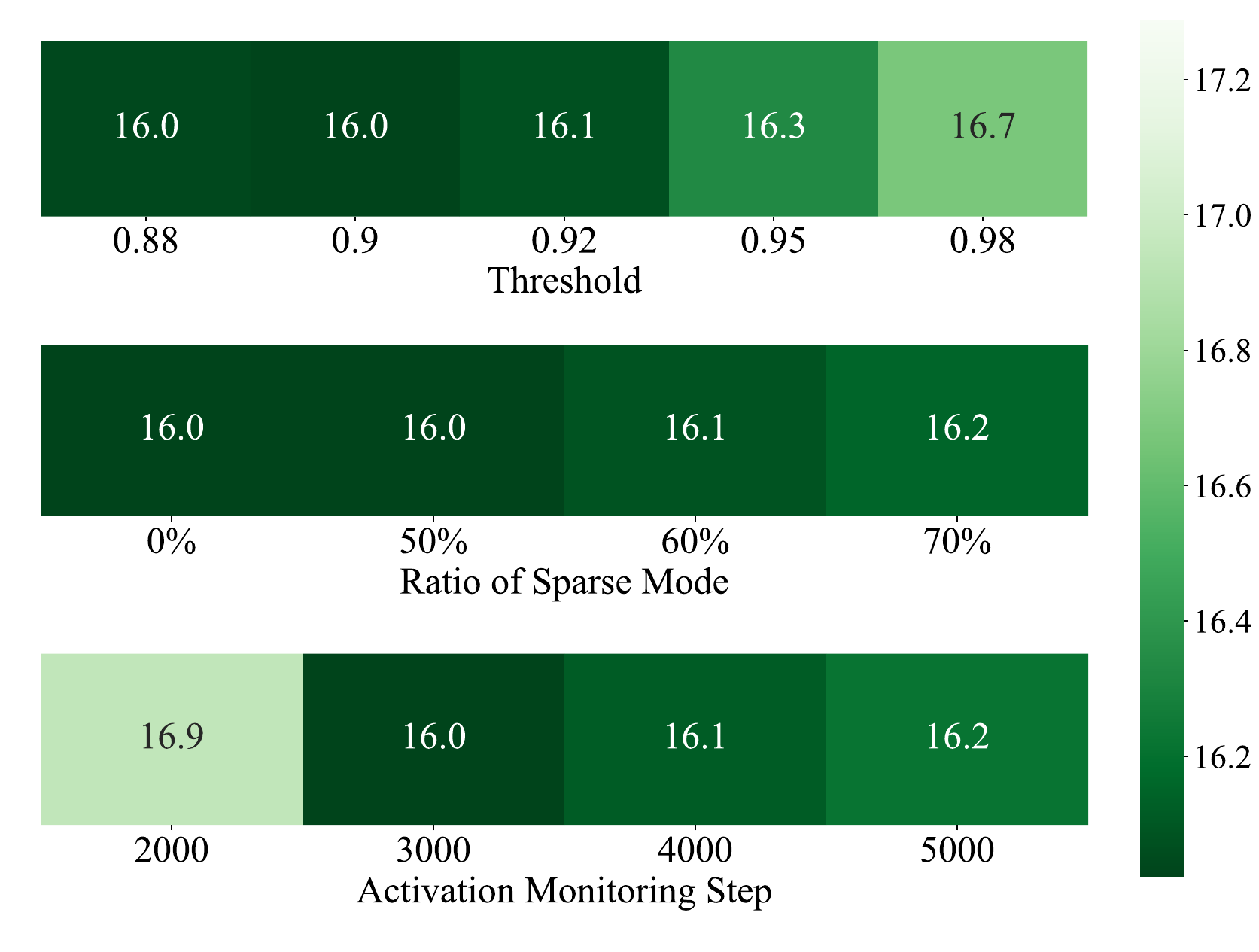}
\caption{Effect of SSD hyperparameters on perplexity. The model architecture is GPT.}
\label{fig:hyper}
\end{figure}

\section{Hyperparameter Analysis}

In this section, we evaluate the effect of the conversion threshold $\tau$, the ratio of the sparse mode $r$, and the steps of monitoring the activation pattern on the performance of GPT with SSD.

The results are shown in Figure~\ref{fig:hyper}.
From this figure, we have three observations.
(1) The conversion threshold $\tau$ cannot be too large, e.g., $0.98$, which will lead to a significant performance drop.
The reason may be that a large $\tau$ will result in a long continuous period of sparse training, which may lead to the overfitting of the SMoE mode and affect the utilization of model capacity.
(2) The ratio of the sparse mode $r$ has little effect on the performance of SSD, which suggests that we can further increase the ratio to achieve a higher speedup.
For example, a sparse ratio of $70\%$ can achieve a FLOPs speedup of $1.63\times$ with a comparable perplexity to dense training.
(3) The steps between two activation pattern monitoring cannot be too small, which may lead to frequent switching and unstable training, e.g., $2000$ steps.
In the future, it is worth exploring the adaptive calculation steps to avoid this issue.

\section{Discussion on GLU Models}

While our work focuses on ReLU-based FFNs, it is worth discussing the applicability of our method to Gated Linear Units (GLU)~\cite{GLU} models, which are widely used in current LLMs~\cite{llama}.
One potential approach would be to modify our clustering algorithm to account for the gating mechanism in GLUs. 
Instead of simply clustering the activation weight matrix, we could jointly cluster the activation and gate weight matrices, treating them as a unified representation of the feedforward unit's behavior. 
This would allow our method to identify sparse modes that capture the intrinsic computational patterns emergent from the interplay between activations and gates.

It is also worth noting that while GLUs introduce additional nonlinearities, they still maintain a level of sparsity in their output representations due to the gating mechanism. As shown in~\citet{relu2}, LLaMA with SwiGLU also exhibits sparse activation phenomenon.

\end{document}